\RequirePackage{snapshot}
\documentclass[letterpaper, 10 pt, conference]{ieeeconf}  %
\IEEEoverridecommandlockouts                              %
\usepackage{epsfig}
\usepackage{booktabs, multicol, multirow}
\usepackage[export]{adjustbox}
\usepackage{amsfonts,amsmath,amssymb,mathtools,bm,bbm}
\usepackage{booktabs}
\usepackage[table]{xcolor}    %
\usepackage[font=footnotesize]{caption}
\newcolumntype{P}[1]{>{\centering\arraybackslash}p{#1}}
\newcolumntype{M}[1]{>{\centering\arraybackslash}m{#1}}

\usepackage{pifont}

\usepackage{algorithm}
\usepackage{algpseudocode}
\makeatletter
\let\OldStatex\Statex
\renewcommand{\Statex}[1][0]{%
  \setlength\@tempdima{\algorithmicindent}%
  \OldStatex\hskip\dimexpr#1\@tempdima\relax}
\makeatother

\algnewcommand\AND{~\textbf{and}~}
\algnewcommand\OR{~\textbf{or}~}
\algnewcommand\CONTINUE{~\textbf{continue}~}

\makeatletter
\algnewcommand{\LineComment}[1]{\Statex \hskip\ALG@thistlm \(\triangleright\)
  #1}
\makeatother

\usepackage{flushend}

\definecolor{fullred}{rgb}{0.85,.0,.1} 
\definecolor{navyblue}{rgb}{.0,.0,.5}
\definecolor{bleudefrance}{rgb}{0.19, 0.55, 0.91}
\definecolor{bluegray}{rgb}{0.18, 0.36, 0.6}
\definecolor{lightgray}{rgb}{0.4, 0.4, 0.4}

\makeatletter
\let\NAT@parse\undefined
\makeatother

\usepackage[numbers,sort]{natbib}
\renewcommand\cite{\citep}
\newcommand{\BIBSTYLE}{abbrvnat}

\newcommand{\vx}{\bm{x}}

\newcommand{\Bt}{\mathbf{t}}

\newcommand{\Bz}{\mathbf{z}}

\newcommand{\BR}{\mathbf{R}}

\newcommand{\Kc}{\mathcal{K}}

\newcommand{\Ic}{\mathcal{I}}
\newcommand{\Xc}{\mathcal{X}}
\newcommand{\Lc}{\mathcal{L}}
\newcommand{\Oc}{\mathcal{O}}

\newcommand{\pnorm}[2]{\left\lVert{#1}\right\rVert_{#2}}

\newcommand{\reals}{\mathbb{R}}

\newcommand{\expnumber}[2]{{#1}\mathrm{e}{#2}}

\newcommand{\xsub}[1]{%
  \mbox{\scriptsize\begin{tabular}{@{}c@{}}#1\end{tabular}}%
  }
\newcommand\underbracewrap[2]{\underbrace{#1}_{\xsub{#2}}}

\newcommand*{\myfiguretitle}[2]{\fontfamily{phv}\fontsize{#2}{#2}\selectfont
#1}

\newcommand*\mycaption[2]{\caption[#1]{\textbf{#1}~$\blacktriangleright$~#2}}

\usepackage[pagebackref=true,breaklinks=true,letterpaper=true,colorlinks,bookmarks=false,urlcolor=navyblue]{hyperref}

\begin{document}

\title{\LARGE \bf Self-Supervised Visual Place Recognition Learning \\in
  Mobile Robots}

\author{Sudeep Pillai and John J. Leonard\\
CSAIL, MIT\\
{\tt\small\{\href{mailto:spillai@csail.mit.edu}{spillai},
  \href{mailto:jleonard@csail.mit.edu}{jleonard}\}@csail.mit.edu}
}

\maketitle
\thispagestyle{empty}

\newcommand{\SECPROJECTDIR}{tex/..}
\newcommand{\SECDIR}{tex}
\newcommand{\seclabel}[1]{\label{\SECPROJECTDIR:#1}} %
\newcommand{\secref}[1]{\ref{\SECPROJECTDIR:#1}} %

\begin{abstract}
Place recognition is a critical component in robot navigation that enables
it to re-establish previously visited locations, and simultaneously use
this information to correct the drift incurred in its dead-reckoned
estimate. In this work, we develop a \textit{self-supervised} approach
to place recognition in robots. The task of visual loop-closure
identification is cast as a metric learning problem, where the labels
for positive and negative examples of loop-closures can be
\textit{bootstrapped} using a GPS-aided navigation solution
that the robot already uses. By leveraging the synchronization between
sensors, we show that we are able to learn an appropriate distance
metric for arbitrary real-valued image descriptors (including
state-of-the-art CNN models), that is
specifically geared for visual place recognition in mobile robots. Furthermore, we show
that the newly learned embedding can be particularly powerful in
disambiguating visual scenes for the task of vision-based loop-closure
identification in mobile robots.
\end{abstract}

\section{Introduction}\seclabel{sec:introduction} Place recognition
for mobile robots is a long-studied topic~\cite{lowry2016visual} due
to the far-reaching impact it will have in enabling fully-autonomous
systems in the near future. State-of-the-art methods for
place-recognition today use hand-engineered image feature descriptors
and matching techniques to implement their vision-based loop-closure
mechanisms. While these model-based algorithms have enabled significant advances in
mobile robot navigation, they are still limited in their ability to
learn from new experiences and adapt accordingly. We envision robots
to be able to learn from their previous experiences and continuously
tune their internal model representations in order to achieve improved
task-performance and model efficiency. With these considerations in
mind, we introduce a bootstrapped mechanism to learn and fine-tune the
model performance of vision-based loop-closure recognition systems in
mobile robots.

With a growing set of experiences that a robot logs today, we
recognize the need for \textit{fully automatic solutions} for
experience-based task learning and model refinement. Inspired by
NetVLAD~\cite{arandjelovic2016netvlad}, we cast the problem of place
recognition in mobile robots as a \textit{self-supervised} metric
learning problem. Most previous
works~\cite{milford2013vision,sunderhauf2015place,lowry2016visual} use
hand-engineered image descriptors or pre-trained Convolutional Neural
Network architectures~\cite{krizhevsky2012imagenet,zhou2014object} to
describe an image for classification or matching. All these methods,
in some way or the other, require a hand-engineered \textit{metric}
for matching the visual descriptors extracted. The choice of feature
extraction needs to be tightly coupled with the right distance metric
in order to retrieve similar objects appropriately. This adds yet
another level of complexity in designing and tuning reliable systems
that are fault tolerant and robust to operating in varying appearance
regimes. Furthermore, these approaches do not provide a mechanism to
optimize for specific appearance regimes (e.g. learn to ignore
fog/rain in those specific conditions). We envision that the distance
metric for these feature descriptors be
learned~\cite{chen2015distance,chen2017deep} from experience, and that
it should be done in a bootstrapped
manner~\cite{kuipers2002bootstrap}. Furthermore, we would prefer that
the features describing the same place to be repeatably embedded close
to each other in some high-dimensional space, with the distances
between them to be \textit{well-calibrated}. With this self-supervised
mechanism, we expect robots to be able to quickly adapt to the visual
appearance regimes it typically sees, and reliably perform visual
place recognition as it gathers more experience.

\begin{figure}[!t]
  \centering 
  \includegraphics[width=\columnwidth]{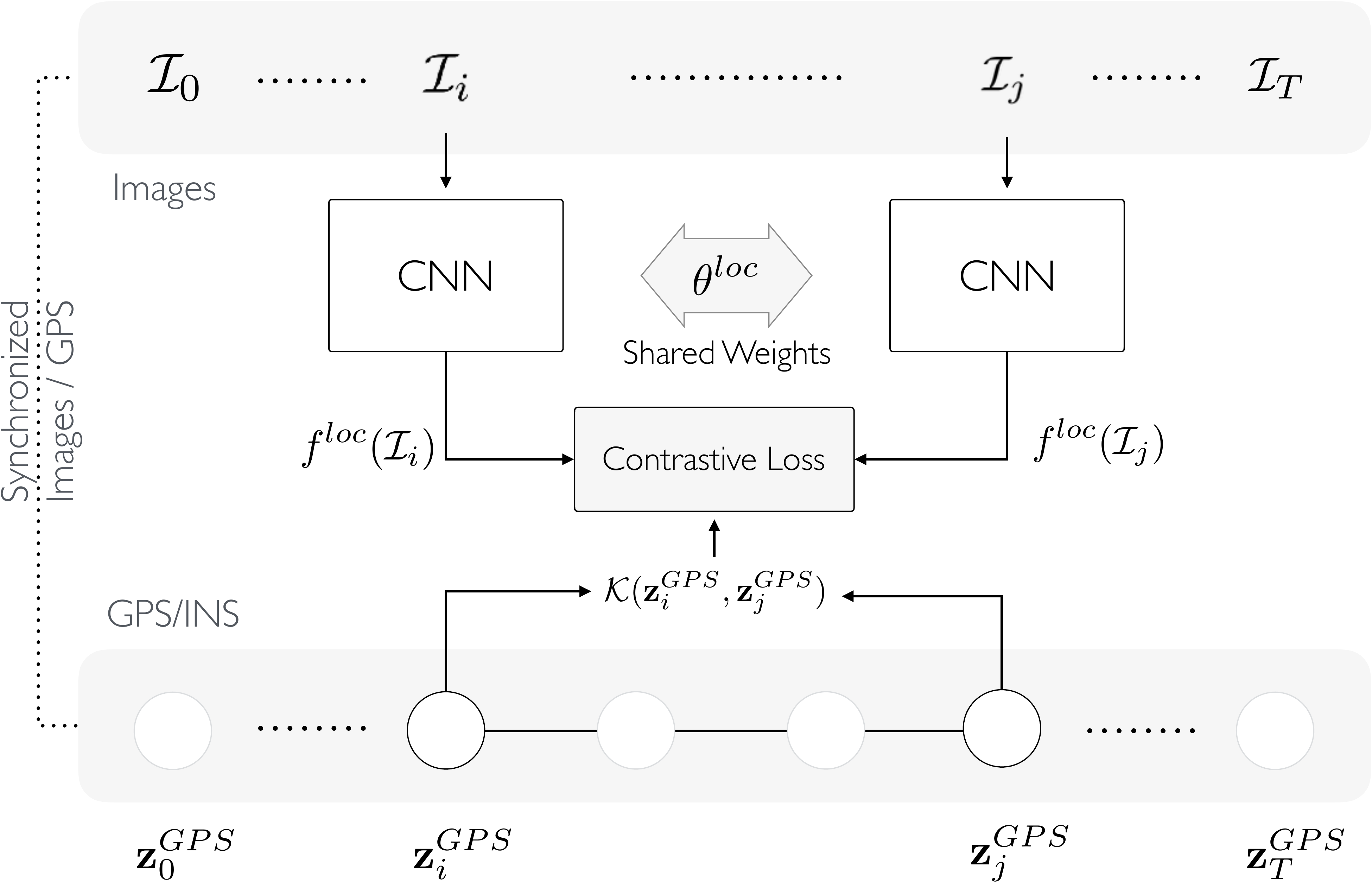}
  \mycaption{Self-Supervised Metric Learning for Localization}{
  The illustration of our proposed self-supervised Siamese Net
  architecture. The model bootstraps synchronized cross-modal information (Images and GPS) in
  order to learn an appropriate similarity metric
  between pairs of images in an embedded space, that implicitly learns
  to predict the loop-closure detection task. The key idea is the ability
  to sample and train our model on positive and negative pairs of examples of similar and
  dissimilar places by taking advantage of corresponding GPS location
  information. }
\seclabel{fig:self-supervised-learning-localization-illustration}
\vspace{-6mm}
\end{figure}

\section{Related Work}\seclabel{sec:related-work} Visual place
recognition in the context of vision-based navigation is a well
studied problem. In order to identify previously visited locations the
system need to be able to extract salient cues from an image that
describes the content contained within it. Typically, the same place
may be significantly different from its previous appearance due to
factors such as variation in lighting (e.g. sunny, cloudy,
rainy etc), observed viewpoint (e.g. viewing from opposite directions,
viewing from significantly different vantage points), or even
perceptual aliasing (e.g. facing and seeing a brick-wall
elsewhere). These properties make it challenging to
hand-engineer solutions that robustly operate in a wide range of
scenarios.

\textbf{Local and Global methods}~~Some of the earliest forms of
visual place recognition entailed directly observing pixel intensities
in the image and measuring their correlation. In order to be invariant
to viewpoint changes, subsequent
works~\cite{kovsecka2005global,mei2010closing,cummins2011appearance,sunderhauf2011brief,furgale2010visual,churchill2012practice}
proposed using low-level \textit{local} and \textit{invariant} feature
descriptors. These
descriptions are aggregated into a single high-dimensional feature
vector via Bag-of-Visual-Words
(BoVW)~\cite{philbin2007object},
VLAD~\cite{jegou2010aggregating} or Fisher
Vectors~\cite{jegou2012aggregating}. Other
works~\cite{sunderhauf2013we,milford2013vision,sunderhauf2011brief}
directly modeled whole-image statistics and hand-engineered
\textit{global} descriptors such as
GIST~\cite{oliva2006building} to determine an
appropriate feature representation for an image.

\textbf{Sequence-based, Time-based or Context-based methods}~~While
image-level feature descriptions are convenient in matching, it
becomes less reliable due to perceptual aliasing, or low saliency
in images. These concerns led further
work~\cite{milford2012seqslam, galveztro12, lynen2014placeless,
maddern2012cat} in matching whole sequences of consecutive images that
effectively describes a place. In SeqSLAM, the
authors~\cite{milford2012seqslam} identify potential loop closures by
casting it as a sequence alignment problem, while
others~\cite{galveztro12} rely on temporal
consistency checks across long image sequences in order to robustly
propose loop closures.~\citet{mei2010closing} finds cliques in the
pose graph to extract place descriptions. \citet{lynen2014placeless}
proposed a placeless-place recognition scheme where they match
features on the level of individual descriptors. By identifying
high-density regions in the distance matrix computed from feature
descriptions extracted across a large sequence of images, the system
could propose swaths of potentially matching places.

\textbf{Learning-based methods}~~In one of the earliest works in
learning-based methods~\citet{kuipers2002bootstrap} proposed a
mechanism to identify distinctive features in a location relative to
those in other nearby locations. In
FABMAP~\cite{cummins2011appearance}, the authors
approximate the joint probability distribution over the
bag-of-visual-words data via the Chow-Liu tree decomposition to
develop an information-theoretic measure for
place-recognition. Through this model, one can sample from the
conditional distribution of visual word occurrence, in order to
appropriately weight the likelihood of having seen identical visual
words before. This reduces the overall rate of false positives,
thereby significantly increasing precision of the system. Another work
from~\citet{latif2013robust} re-cast place-recognition as a sparse
convex $L_1$ minimization problem with efficient homotopy methods that
enable robust loop-closure hypothesis. In similar light,
experience-based learning
methods~\cite{churchill2012practice,furgale2010visual} take advantage of
the robot's previous experiences to learn the set of features to
match, incrementally adding more details to the description of a place
if an existing description is insufficient to match a known place.

\textbf{Deep Learning methods}~~Recently, the advancements in
Convolutional Neural Network (CNN)
Architectures~\cite{krizhevsky2012imagenet,simonyan2014very} have
drastically changed the landscape of algorithms used in vision-based
recognition tasks. Their adoption in vision-based place recognition
for robots~\cite{sunderhauf2015place,chen2017deep} have recently shown
promising results. However, most domain-specific tasks require further
model fine-tuning of these large-scale networks in order to perform
reliably well. Despite the ready
availability of training models and weights, we foresee the data
collection and its supervision being a predominant source of friction
for fine-tuning models for tasks such as place recognition in
robots. Due to the rich amount of cross-modal information that robots
typically collect, we expect to \textit{self-supervise} tasks such as
place recognition by fine-tuning existing CNN models with the
experience they have accumulated. To this end, we fine-tune these
feature representations specifically for the task of loop-closure
recognition and show significant improvements in the precision-recall
performance.

\begin{figure*}[!t]
  \centering 
  \colorbox{white}{\renewcommand{\arraystretch}{0.6} %
    {\setlength{\tabcolsep}{0.3mm}
      \setlength\fboxsep{0pt}
      \setlength\fboxrule{0.5pt}
      \begin{tabular}{ccccccc}
        \hspace{1mm}\includegraphics[height=0.9in]{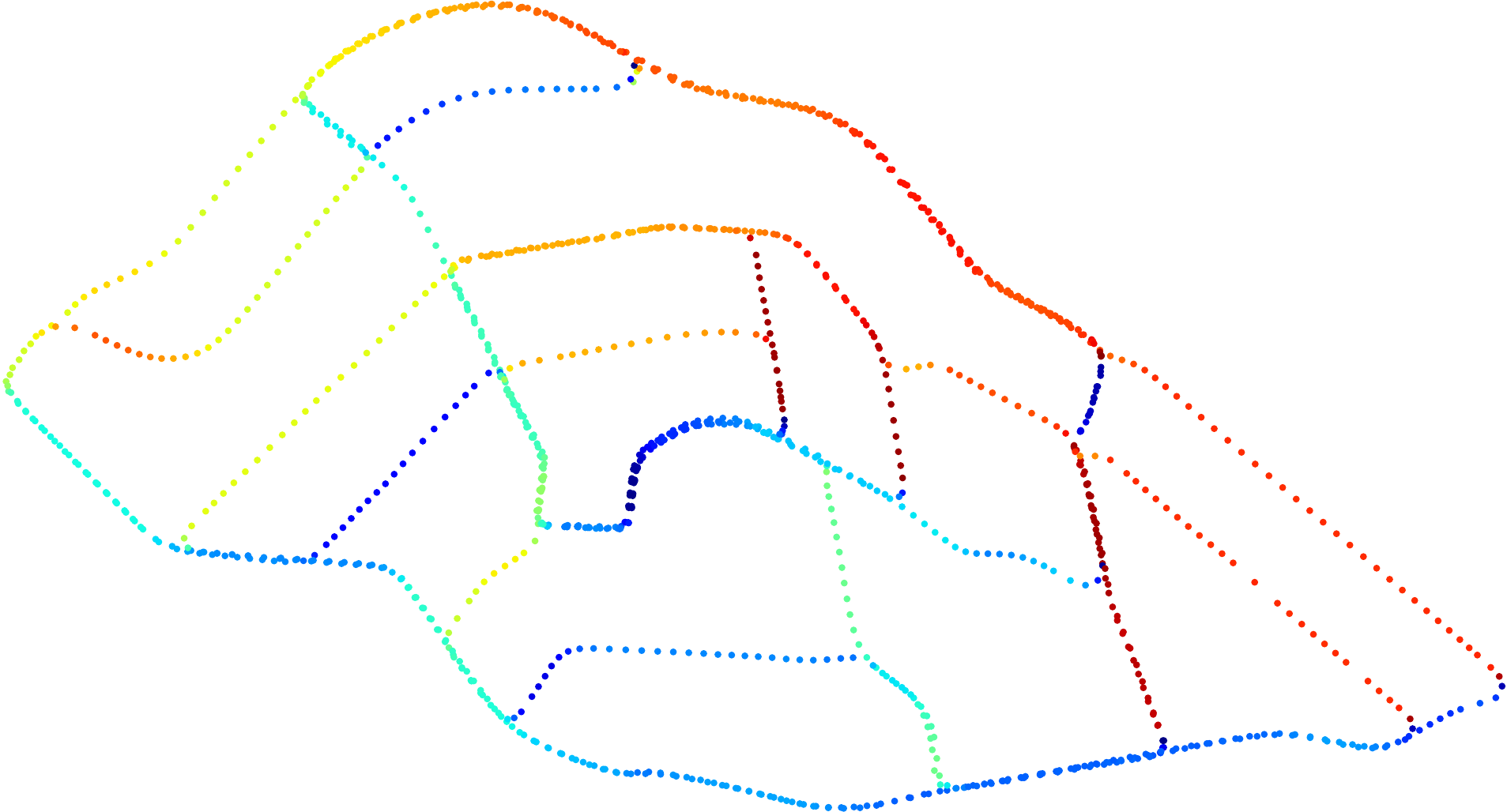}\hspace{1mm}
        &\includegraphics[fbox,height=0.95in]{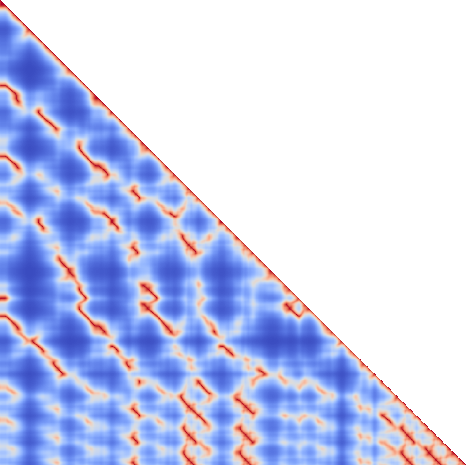}
        &\includegraphics[fbox,height=0.95in]{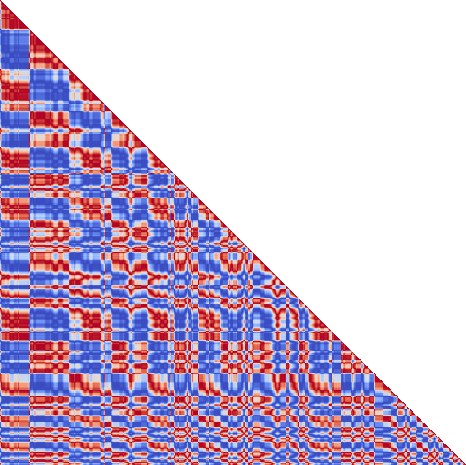}
        &\includegraphics[fbox,height=0.95in]{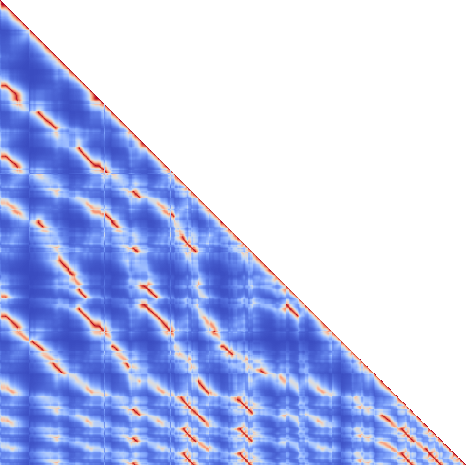}
        &\includegraphics[fbox,height=0.95in]{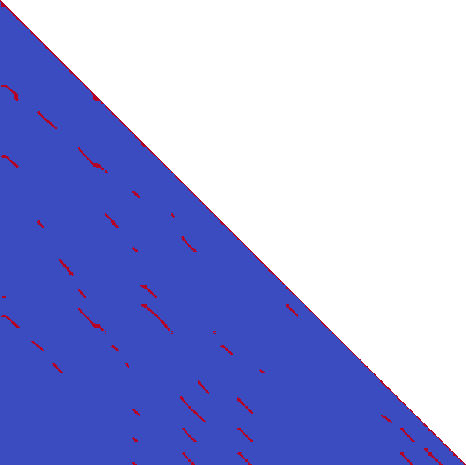}
        &\includegraphics[fbox,height=0.95in]{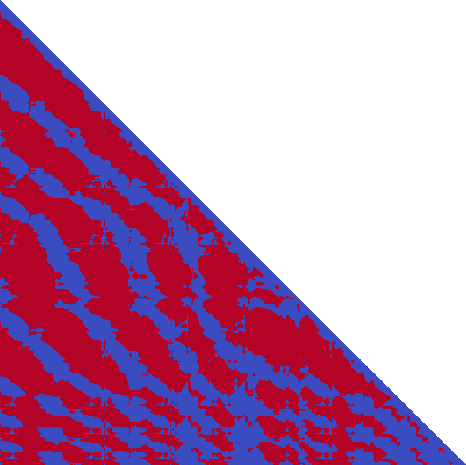}
          &\begin{tabular}[b]{c}
           {\scriptsize 1.0}\\
           \includegraphics[frame,width=0.75in,angle=90]{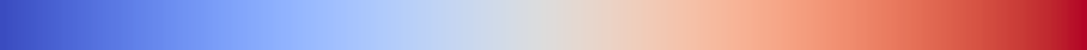}\\
             {\scriptsize 0.0}
           \end{tabular}\\
        {\scriptsize \textbf{St. Lucia Dataset}} 
        
        &{\scriptsize \textbf{Translation ($\mathbf{t}$)}} & {\scriptsize
                                                  \textbf{Rotation ($\mathbf{R}$)}}
        &{\scriptsize \textbf{Rot. \& Trans. ($\mathbf{Rt}$)}}
        &{\scriptsize \textbf{Positive Labels}}          
        &{\scriptsize \textbf{Negative Labels}}&
        
      \end{tabular}}}
  \mycaption{Bootstrapped learning using cross-modal information}{
  An illustration of the vehicle path traversed in the St. Lucia dataset
    (\texttt{100909\_1210}) with synchronized Image and GPS
    measurements. The colors correspond to the vehicle bearing angle
    (Rotation $\BR$) inferred from the sequential GPS
    measurements. The self-similarity matrix determined from
    the translation ($\Bt$), rotation ($\BR$) and their combination
    ($\BR\Bt$) on the St. Lucia Dataset using the assumed ground-truth
    GPS measurements. Each row and column in the self-similarity matrix corresponds to
    keyframes sampled from the dataset as described in
    Section~\secref{subsec:self-supervised-dataset-generation}. The
    sampling scheme ensures a time-invariant (aligned) representation
    where loop-closures appear as off-diagonal entries that are a
    fixed-offset from the current sequence (main-diagonal). We use a
    Gaussian kernel (Equation~\secref{eq:gps-similarity}) to describe
    the similarity between keyframes and sample positive/negative
    samples from the combined $\BR\Bt$ similarity matrix. The $\Kc$ kernel computed
  in Equation~\secref{eq:gps-similarity} is used to ``supervise'' the
  sampling procedure. \textbf{Positive Labels:} Samples whose
  kernel $\Kc(\Bz^{GPS}, \Bz'^{GPS})$ evaluates to higher than
  $\tau_{p}^{\BR\Bt}$ are considered as positive samples (in
  red). \textbf{Negative Labels:} Samples whose kernel
  $\Kc(\Bz^{GPS}, \Bz'^{GPS})$ evaluates to lower than
  $\tau_{n}^{\BR\Bt}$ are consider as negative examples (in
  red).}
  \seclabel{fig:self-supervised-sampling}
  \vspace{-4mm}
\end{figure*}

\section{Background: Metric Learning}%
\seclabel{sec:metric-learning-background}

In this work we rely on metric learning to learn an appropriate metric
for the task of place recognition in mobile robots. The problem of metric
learning was first introduced as \textit{Mahalanobis metric learning}
in~\cite{xing2003distance}, and subsequently
explored~\cite{kulis2013metric} with various dimensionality-reduction,
information-theoretic and geometric lenses. More abstractly, metric learning seeks to learn a non-linear mapping
$f(\cdot;\theta) : \reals^n \to \reals^m$ that takes in input data pairs
$(\vx_i, \vx_j) \in \reals^n$, where the Euclidean distance in the new
target space $\pnorm{f(\vx_i;\theta)-f(\vx_j;\theta)}{2}$ is an
approximate measure of \textit{semantic} distance in the original
space $\reals^n$. Unlike in the supervised learning paradigm where the
loss function is evaluated over individual samples, here, we consider
the loss over pairs of samples $\Xc = \Xc_S \cup \Xc_D$. We define
sets of similar and dissimilar paired examples $\Xc_S$, and $\Xc_D$
respectively as follows
\begin{align}
  \seclabel{eq:positive-negative-labels}
  \Xc_S &:= \{(\vx_q,\vx_s) \mid \vx_q \; \text{and} \; \vx_s \; \text{are in the \textit{same} class} \}\\
  \Xc_D &:= \{ (\vx_q,\vx_d) \mid \vx_q \; \text{and} \; \vx_d \; \text{are in \textit{different} classes} \}
\end{align}
and define an appropriate loss function that captures the
aforementioned properties. 

\textbf{Contrastive Loss} The contrastive
loss introduced by~\citet{chopra2005learning} optimizes the distances between
positive pairs $(\vx_q, \vx_s)$ such that they are drawn closer to each
other, while preserving the distances between negative pairs $(\vx_q,
\vx_d)$ at or above a fixed margin $\alpha$ from each other. Intuitively, the overall
loss is expressed as the sum
of two terms with $y$ being the indicator variable in identifying positive
examples from negative ones, 
\begin{align}
  \seclabel{eq:contrastive-loss-minimal}
  \Lc(\theta) = \sum_{(\vx_i,\vx_j) \in \Xc}
  yD_{ij}^2 + (1-y)\Big[\alpha - D_{ij}\Big]_{+}^2
\end{align}\vspace{-4mm}
\begin{align}
  \seclabel{eq:semantic-distance-theirs}
  \text{where}~~D_{ij} &= \pnorm{f(\vx_i;\theta) - f(\vx_j;\theta)}{2}^2\\
  \seclabel{eq:semantic-indicator-theirs}
  \text{and}~~y &= \begin{cases}
    1 & \text{if } (\vx_i,\vx_j) \in \Xc_S, \\
    0 & \text{if } (\vx_i,\vx_j) \in \Xc_D \\
  \end{cases}
\end{align}

\textbf{Training with Siamese Networks}~~Learning is then typically
performed with a Siamese
architecture~\cite{bromley1994signature,chopra2005learning},
consisting of two parallel networks $f(\vx;\theta)$ that share weights
$\theta$ amongst each other. The
contrastive loss is then defined between the two parallel networks
$f(\vx_i;\theta)$ and $f(\vx_j;\theta)$ given by
Equation~\secref{eq:contrastive-loss-minimal}. The scalar output loss computed
from batches of similar and dissimilar samples are then used to update
the parameters of the Siamese network $\theta$ via Stochastic Gradient
Descent (SGD). Typically, batches of positive and negative samples are
provided in alternating fashion during training.

\section{Self-Supervised Metric Learning \\for Place
  Recognition}\seclabel{sec:procedure}

\begin{figure*}[!th]
  \centering 
  \colorbox{white}{\renewcommand{\arraystretch}{1.0} %
    {\setlength{\tabcolsep}{0.3mm}
      \setlength\fboxsep{0pt}
      \setlength\fboxrule{0.5pt}
      \begin{tabular}{cccccc}
        \includegraphics[fbox,height=1.3in]{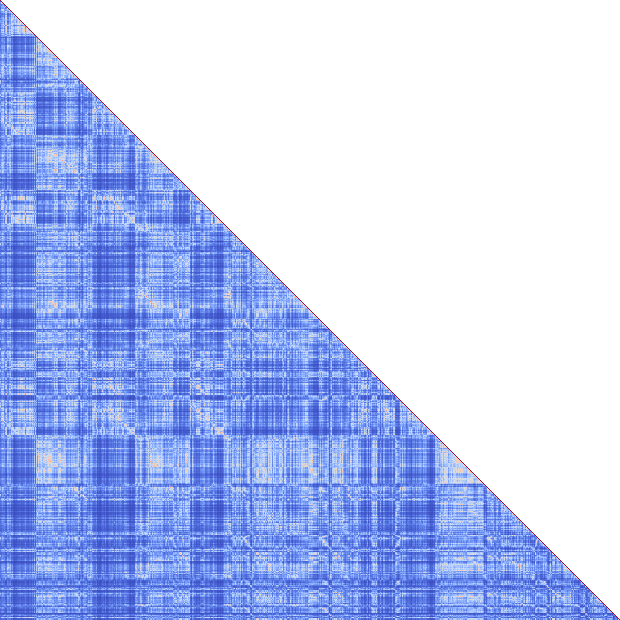}
        &\includegraphics[fbox,height=1.3in]{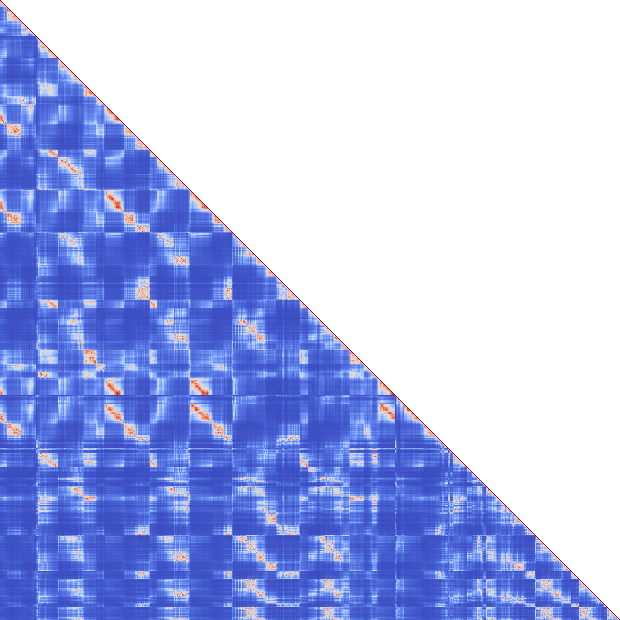}
        &\includegraphics[fbox,height=1.3in]{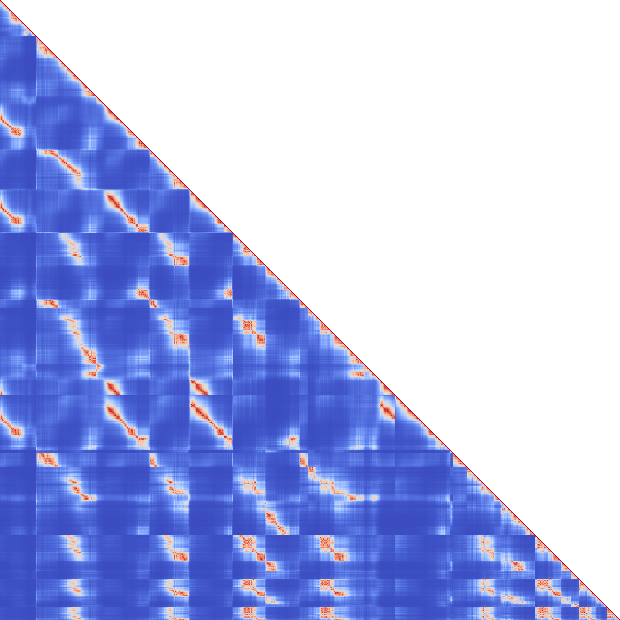}
        &\includegraphics[fbox,height=1.3in]{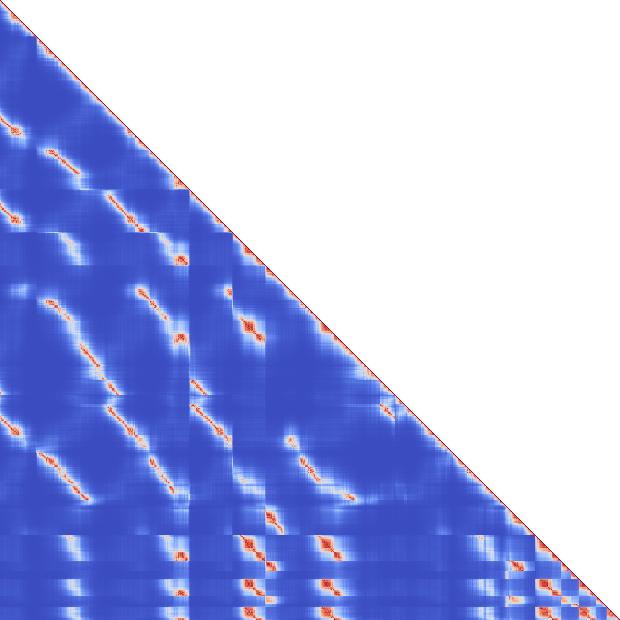}
        &\includegraphics[fbox,height=1.3in]{results/st-lucia-100909_1210/pose_Rt_cmatrix.png}
        &\begin{tabular}[b]{c}
           {\scriptsize 1.0}\\
           \includegraphics[frame,width=1in,angle=90]{graphics/colormaps/coolwarm.pdf}\\
           {\scriptsize 0.0}
         \end{tabular}
        \\{\scriptsize \textbf{Epoch 0}}
        & {\scriptsize \textbf{Epoch 30}}
        &{\scriptsize \textbf{Epoch 180}}
        &{\parbox{0.8in}{\centering \scriptsize \textbf{Learned Similarity\\ (Final epoch)}}}
        & {\parbox{1in}{\centering \scriptsize \textbf{Target
          GPS Similarity\\ (From
        Figure~\secref{fig:self-supervised-sampling})}}}& 
      \end{tabular}}}
  \mycaption{Self-Supervised learning of a visual-similarity
      metric}{ An illustration of the similarity matrix at various stages of
    training. At \textit{Epoch 0}, the distances between features
    extracted at identical locations are not well-calibrated requiring
    hand-tuned metrics for reliable matching. With more positive and
    negative training examples, the model at \textit{Epoch 30} has learned
    to draw positive features closer together (strong red off-diagonal
    sequences indicating loop-closures), while pushing negative features
    farther apart (strong blue background). This trend continues with
    \textit{Epoch 180} where the loop-closures start to look well-defined,
    while the background is consistently blue indicating a reduced
    likelihood for false-positives. Comparison of the learned visual-similarity
    metric against the target or ground truth similarity metric (obtained
    by determining overlapping frustums using GPS measurements). As
    expected, the distances in the learned model tend to be
    \textit{well-calibrated} enabling strong precision-recall
    performance. Furthermore, the model can be qualitatively validated
    when the learned similarity matrix starts to closely resemble the
    target similarity matrix (comparing columns 2 and 3 in the figure).}
\seclabel{fig:metric-learning-epochs-1} \vspace{-4mm}
\end{figure*}

\subsection{Self-supervised dataset generation}
\seclabel{subsec:self-supervised-dataset-generation}

Multi-camera systems and navigation modules have more-or-less become
ubiquitous in modern autonomous systems today. Typical systems log this
sensory information in an asynchronous manner, providing a treasure of
cross-modal information that can be readily used for transfer learning
purposes. Here, we focus on the task of vision-based place recognition
via a forward-looking camera, by leveraging synchronized information
collected via standard navigation modules (GPS/IMU, INS etc.).

\textbf{Sensor Synchronization}~~In order to formalize the notation
used in the following sections, we shall refer to
$(\Ic_t,\Bz^{GPS}_t)$ as the \textit{synchronized} tuple of camera
image $\Ic$, and GPS measurement $\Bz^{GPS}$, captured at
approximately the same time $t$. In typical systems however, these
sensor measurements are captured in an asynchronous manner, and the
synchronization needs to be carried out carefully in order to ensure
clean and reliable measurements for the bootstrapping procedure. It is
important to note that for the specific task of place recognition,
$\Bz$ can be also be sourced from external sensors such as
inertial-navigation systems (INS), or even recovered from a GPS-aided
SLAM solution.

\textbf{Keyframe Sub-sampling}~~While we could consider the full set
of synchronized image-GPS pairs, it may be sufficient to learn only
from a diverse set of viewpoints. We expect that learning from this
strictly smaller, yet diverse set, can substantially
speed up the training process while being able to achieve the same
performance characteristics when trained with the original
dataset. While it is unclear what this sampling function may look like
for image descriptions, we can easily provide this measure to
determine a diverse set of GPS measurements. We incorporate this via a
standard keyframe-selection strategy where the poses are sampled from
a continuous stream whenever the relative pose has exceeded a certain
translational or rotational threshold from its previously established
keyframe. We set these translational and rotational thresholds to 5m,
and $\frac{\pi}{6}$ radians respectively to allow for efficient
sampling of diverse keyframes.

\textbf{Keyframe Similarity}~~The self-supervision is enabled by
defining a viewing frustum that applies to both the navigation-view $\Bz_t$
and the image-view $\Ic$. We define a Gaussian similarity kernel $\Kc$ between two instances
of GPS measurements $\Bz_i^{GPS}$ and $\Bz_j^{GPS}$  given by
$\Kc(\Bz_i^{GPS}, \Bz_j^{GPS})$, (or $\Kc_{ij}$ in short):
\newcommand{\gaussianGPStranslation}{\exp(-\gamma^{\Bt}\pnorm{\Bz_i^{\Bt} - \Bz_j^{\Bt}}{2}^2)}
\newcommand{\gaussianGPSrotation}{\exp(-\gamma^{\BR}\pnorm{\Bz_i^{\BR} \ominus \Bz_j^{\BR}}{2}^2)}
\begin{align}
  \seclabel{eq:gps-similarity}
  \Kc_{ij} =
  \underbracewrap{\gaussianGPStranslation}{Translation similarity} \cdot
  \underbracewrap{\gaussianGPSrotation}{Rotation similarity}
\end{align}
where $\Bz_i^{\Bt}$ is the GPS translation measured in
metric-coordinates at time $i$, and $\Bz_i^{\BR}$ is the corresponding
rotation or bearing determined from the sequential GPS coordinates for
the particular session (See
Figure~\secref{fig:self-supervised-sampling}). Here, the only
hyper-parameter required is the choice of
the bandwidth parameters
$\gamma^{\BR}$ and $\gamma^{\Bt}$, and generally depends on the
viewing frustum of the camera used. The resulting
self-similarity matrix for the translation (using GPS translation $\Bt$
only), and the rotation (using established bearing $\BR$ only) on a
single session from the St. Lucia Dataset~\cite{glover2010fab} is
illustrated in
Figure~\secref{fig:self-supervised-sampling}.

\textbf{Distance-Weighted Sampling}~~With keyframe based sampling
considerably reducing the dataset for efficient training, we now focus
on sampling positive and negative pairs in order to ensure speedy
convergence of the objective function. We first consider the keyframe
self-similarity matrix between all pairs of keyframes for a given dataset,
and sample positive pairs whose similarity exceeds a specified
threshold $\tau_{p}^{\BR\Bt}$. Similarly, we sample negative pairs
whose similarity is below $\tau_{n}^{\BR\Bt}$. For each of the
positive and negative sets, we further sample uniformly by their
inverse distance in the original feature space
following~\cite{1706.07567}, to encourage faster convergence.

\subsection{Learning an appropriate distance metric for localization}
\seclabel{subsec:distance-learning}

Our proposed self-supervised place recognition architecture is
realized with a Siamese network with an appropriate contrastive
loss~\cite{chopra2005learning} (given by
Equation~\secref{eq:contrastive-loss-ours}). This simultaneously finds
a reduced dimensional metric space where the relative distances
between features in the embedded space are
\textit{well-calibrated}. Here, well-calibrated refers to
the property that negative samples are separated at least by a known
margin $\alpha$, while positive samples are likely to be separated by
a distance less than the margin. Following the terminology in
Section~\secref{sec:metric-learning-background}, we consider tuples
$(\Ic_i, \Bz_i^{GPS}) \in \Xc$ of similar (positive) $\Xc_S \subset
\Xc$ and dissimilar (negative) examples $\Xc_D \subset \Xc$ for
learning an appropriate embedding function
$f^{loc}(\cdot;\theta^{loc})$. Intuitively, we seek to find a
``\textit{semantic measure}'' of distance given by $D(\Ic_i, \Ic_j) =
\pnorm{f^{loc}(\Ic_i; \theta^{loc}) - f^{loc}(\Ic_j; \theta^{loc})}{2}$
in a target space of $\reals^m$ such that they respect the kernel
$\Kc_{ij}$ defined over the space of GPS
measurements as given in
Equation~\secref{eq:gps-similarity}.

Let $(\Ic,\Bz^{GPS}) \in \Xc$ be the input data and $\mathbbm{1}_{G} \in {0,1}$
be the indicator variable representing dissimilar
($\mathbbm{1}_{G}=0$) and similar ($\mathbbm{1}_{G}=1$) pairs of examples
within $\Xc$.  We seek to find a function $f^{loc}(\cdot;\theta^{loc}): \Ic
\mapsto \Phi$ that maps the input image $\Ic$ to an embedding
$\Phi \in \reals^m$ whose distances between similar places
are low, while the distances between dissimilar places are
high. We take advantage of availability of synchronized Image-GPS
measurements $(\Ic, \Bz^{GPS})$ to provide an indicator for place
similarity, thereby rendering this procedure fully automatic or
self-supervised. Re-writing
equation~\secref{eq:contrastive-loss-minimal} for our problem, we get
Equation~\secref{eq:contrastive-loss-ours} where $D(\Ic_i,\Ic_j)$
measures the ``\textit{semantic distance}'' between images
(Equation~\secref{eq:semantic-distance-ours}). 
\begin{align}
  \Lc(\theta^{loc}) &= \sum_{\Xc}~
  \mathbbm{1}_{{\tiny G}} \cdot D_{ij}^2 + 
                (1-\mathbbm{1}_{{\tiny G}}) \cdot \Big[\alpha -
  D_{ij}\Big]_{+}^2
  \seclabel{eq:contrastive-loss-ours}
\end{align}\vspace{-4mm}
\begin{align}
  \seclabel{eq:semantic-distance-ours}
  \text{where}~~D_{ij} &=
  \pnorm{f^{loc}(\Ic_i;\theta^{loc})-f^{loc}(\Ic_j;\theta^{loc})}{2}\\\vspace{4mm}
  \seclabel{eq:semantic-indicator-ours}
  \text{and}~~\mathbbm{1}_{G} &= \begin{cases}
    1 & \text{if}~~\Kc(\Bz^{GPS}_i, \Bz^{GPS}_j) > \tau_{p}^{\BR\Bt} \\
    0 & \text{if}~~\Kc(\Bz^{GPS}_i, \Bz^{GPS}_j) < \tau_{n}^{\BR\Bt} \\
  \end{cases}
\end{align}
For brevity, we omit $\theta^{loc}$ and use
$f^{loc}(\Ic_i)$ instead of the full expression
$f^{loc}(\Ic_i;\theta^{loc})$. We pick the thresholds for $\tau^{\BR\Bt}$ based on a
combination of factors including convergence rate and overall accuracy
of the final learned metric. Nominal values of $\tau_{p}^{\BR\Bt}$ range
from 0.8 to 0.9 that indicate the tightness of the overlap between
viewing frustums of positive examples, with $\tau_{n}^{\BR\Bt}$ for
negative examples set to 0.4.

Figure~\secref{fig:metric-learning-epochs-1} illustrates the visual
self-similarity matrix of the feature embedding at various stages
during the training process on the St. Lucia Dataset
(\texttt{100909\_1210}). At \textit{Epoch 0}, when the feature
embedding is equivalent to the original feature description, it is
hard to disambiguate potential loop-closures due to the \textit{uncalibrated}
nature of the distances. As training progresses, the
positively labeled examples of
loop-closure image pairs are drawn closer together in the embedded space, while
the negative examples are pushed farther from each other. As the training
converges, we start to notice a few characteristics in the learned
embedding that make it especially powerful in identifying
loop-closures: (i) The red diagonal bands in the visual
self-similarity matrix are well-separated from the blue background
indicating that the learned embedding has identified a more separable
function for the purposes of loop-closure recognition; and (ii) The
visual self-similarity matrix starts to resemble the target
self-similarity matrix computed using the GPS measurements (as shown
in Figure~\secref{fig:metric-learning-epochs-1}). Furthermore, the
t-SNE embedding~\cite{maaten2008visualizing} (colorization) of the
learned features extracted at identical locations are strikingly
similar, indicating that the learned feature embedding $f(\cdot;
\theta^{loc})$ has implicitly learned a metric space that is more
appropriate for the task of place-recognition in mobile robots.

\begin{figure}[!t]
  \centering 
  {\renewcommand{\arraystretch}{0.6} %
    {\setlength{\tabcolsep}{0.2mm}
      \begin{tabular}{c}
        \includegraphics[width=0.8\columnwidth]{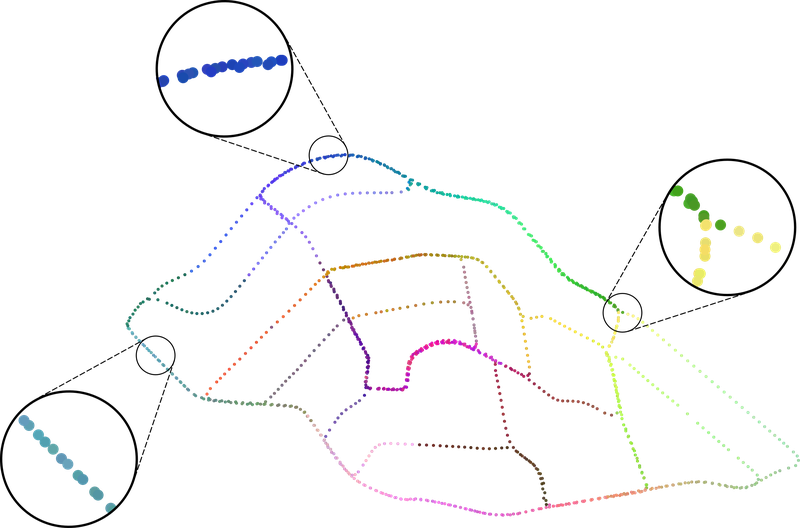}\\
      \end{tabular}}}
  \mycaption{Trajectory with features embedded via T-SNE} { 
    An illustration of the path traversed (\texttt{100909\_1210}) with
    the colors indicating the 3-D t-SNE embedding of the learned features
    $\Phi$ extracted at those corresponding locations. The visual features
    extracted across multiple traversals along the same location are
    consistent, as indicated by their similar color embedding. colors are
    plotted in the RGB colorspace. }
  \seclabel{fig:tsne-trajectory}
  \vspace{-6mm}
\end{figure}

\subsection{Efficient scene indexing, retrieval and
  matching}\seclabel{subsec:scene-retrieval}

One of the critical requirements for place-recognition is to ensure
high recall in loop-closure proposals while maintaining sufficiently
high precision in candidate matches. This however requires
probabilistic interpretability of the matches proposed, with accurate
measures of confidence in order to incorporate these measurements into
the back-end pose graph optimization. Similarities or
distances measured in the image descriptor space are not
well-calibrated, making these measures only amenable to
distance-agnostic matching such as $k$-nearest neighbor
search. Moreover, an indexing and matching scheme such as $k$-nn also
makes it difficult to filter out false positives as the distances
between descriptors in the original embedding space is practically
meaningless. Calibrating distances by learning a new embedding has the
added advantage of avoiding these false positives, while being able to
recover confidence measures for the matches retrieved.

Once feature embedding is learned, and the features $\Phi$ are mapped to an
appropriate target space, we require a mechanism to
insert and query these embedded descriptors from a database. We
use a KD-Tree in order to incrementally insert features into a
balanced tree data structure, thereby enabling $\Oc(\log{}N)$
queries.

\section{Experiments and Results}\seclabel{sec:results}

We evaluate the performance of the proposed self-supervised
localization method on the KITTI~\cite{Geiger2012CVPR} and St. Lucia
Dataset~\cite{glover2010fab}. We compare our approach against the image descriptions obtained
from extracting the activations from several layers in the
Places365-AlexNet pre-trained model~\cite{zhou2016places}
(\textit{conv3},~\textit{conv4},~\textit{conv5},~\textit{pool5},~\textit{fc6},~\textit{fc7}
and \textit{fc8} layers). While we take advantage of the pre-trained
models developed in~\cite{zhou2016places} for the following
experiments, the proposed framework could allow us to learn relevant
task-specific embeddings from any real-valued image-based feature
descriptor.  The implementation details of our proposed method is
described in detail in section~\secref{subsec:implementation-details}.

\subsection{Learned feature embedding characteristics}
\seclabel{subsec:original-learned-metric-compare}

While pre-trained models can be especially powerful image descriptors,
they are typically trained on publicly-available datasets such as the
ImageNet~\cite{ilsvrc15}. that have strikingly different
natural image statistics. Moreover, some of these models are trained
for the purpose of image or place classification. As with most
pre-trained models, we expect some of the descriptive performance of
Convolutional Neural Networks to generalize, especially in its
lower-level layers
(\textit{conv1},~\textit{conv2},~\textit{conv3}). However, the
descriptive capabilities in its mid-level and higher-level layers
(\textit{pool4},~\textit{pool5},~\textit{fc} layers) start to
specialize to the specific data regime and recognition task it is
trained for. This has been addressed quite extensively in the
literature, arguing the need for domain adaptation and fine-tuning
these models on more representative datasets to improve task-specific
performance~\cite{gopalan2011domain,ganin2015unsupervised}.

Similar to previous domain adaptation
works~\cite{gopalan2011domain,ganin2015unsupervised,glorot2011domain},
we are interested in adapting existing models to newer task domains
such as place-recognition with minimal human supervision involved. We
argue for a self-supervised approach to model fine-tuning, and
emphasize the need for a well-calibrated embedding space, where the
features embedded in the new space can provide measures for both
similarity and the corresponding confidence associated in matching.

\textbf{Comparing performance between the original and learned embedding
  space}~~In Figure~\secref{fig:original-learned-comparison-all}, we compare the
precision-recall performance in loop-closure recognition using the
original and learned feature embedding space. For various thresholds
of localization accuracy (20 and 30 meters), our learned embedding shows considerable
performance boost over the pre-trained Places365-AlexNet model. In the figures, we also
illustrate the noticeable drop in performance with the descriptive
capabilities in the higher-level layers
(\textit{fc6},~\textit{fc7},~\textit{fc8}) as compared to the
lower-level layers (\textit{conv3},~\textit{conv4},~\textit{conv5}) in the
Places365-AlexNet model. This is as expected, since the higher layers
in the CNN (\textit{pool5},~\textit{fc6},~\textit{fc7}) are more tailed
to the original classification task they were trained for.

\begin{figure}[!h]
  \centering 
  {\renewcommand{\arraystretch}{0.6} %
    {\setlength{\tabcolsep}{0.2mm}
      \begin{tabular}{cc}
        \hspace{4mm}\myfiguretitle{Precision-Recall comparison at 20m}{7}
        & \hspace{4mm}\myfiguretitle{Precision-Recall comparison at 30m}{7} \\
        \includegraphics[width=0.5\columnwidth]{\SECPROJECTDIR/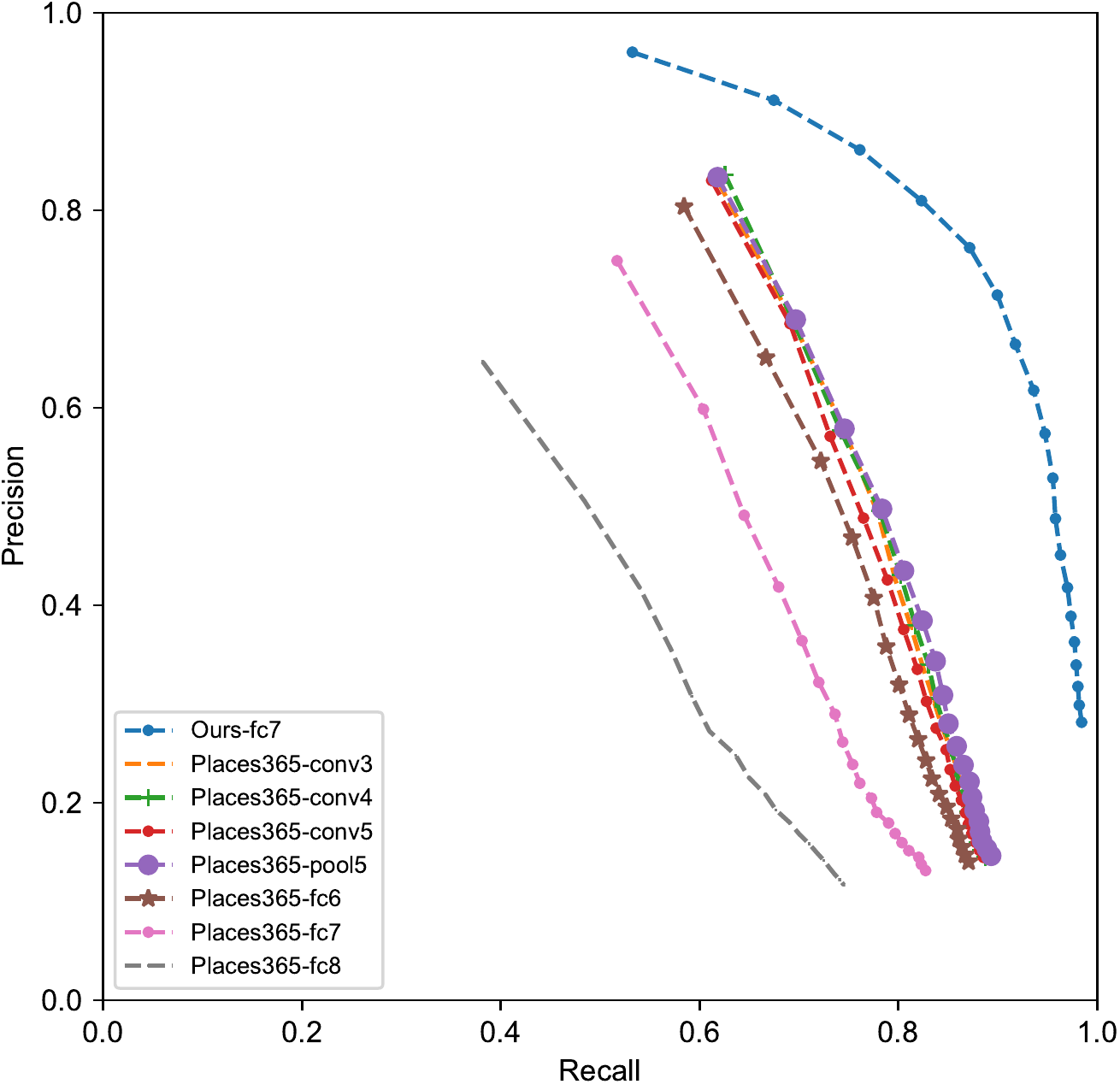}&
        \includegraphics[width=0.5\columnwidth]{\SECPROJECTDIR/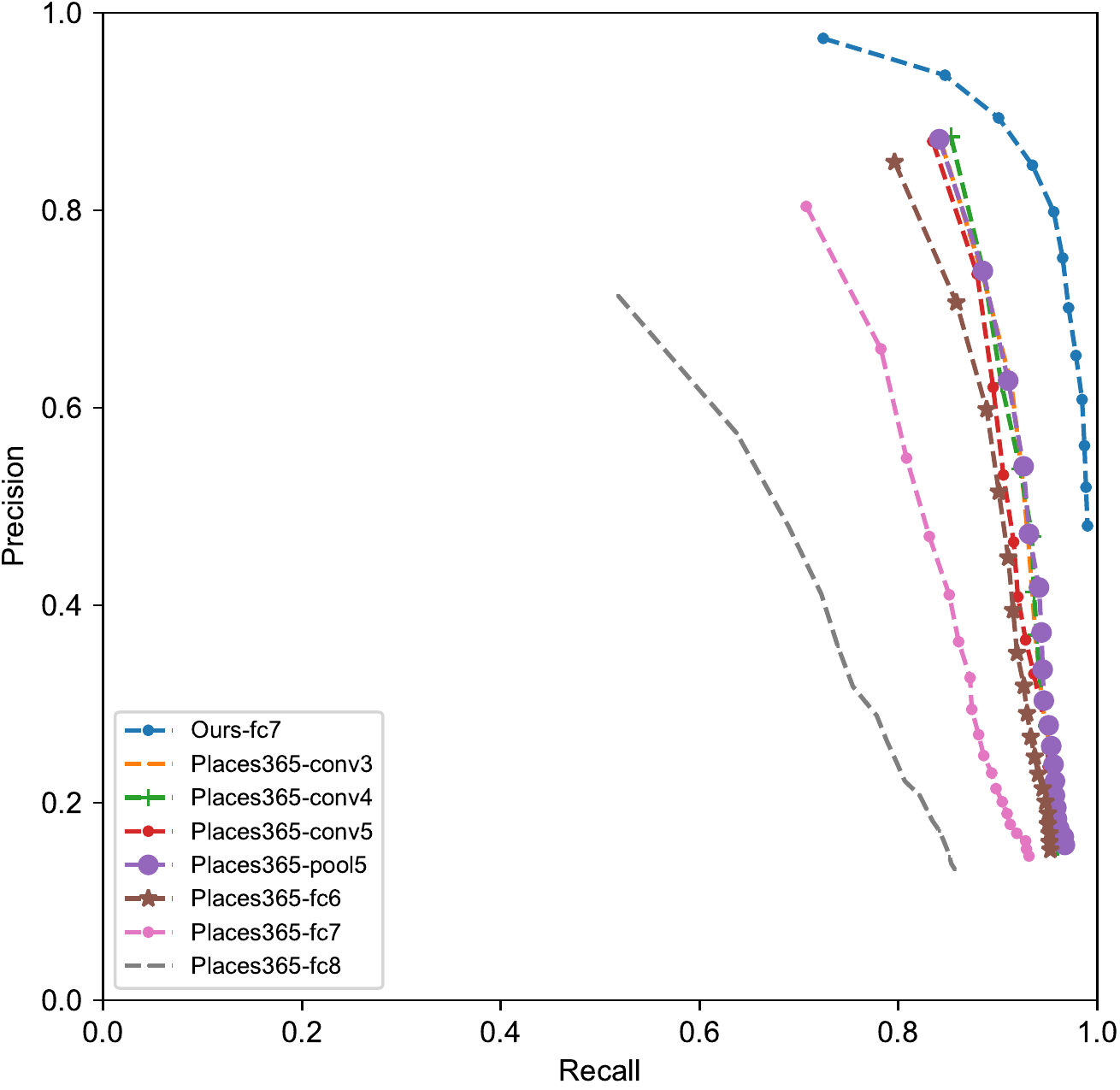}\\
      \end{tabular}}}
  \mycaption{Precision-Recall performance in loop-closure recognition using the 
      original and learned feature embedding space}{ The figures show
    the precision-recall (P-R) performance in loop-closure recognition
    for various feature descriptors using the pre-trained
    Places365-AlexNet model and the learned embedding
    (\textit{Ours-fc7}). Our learned embedding is able to
    significantly outperform the pre-trained Places365-AlexNet model
    for all feature layers, by \textit{self-supervising} the model
    on a more representative dataset.} 
  \seclabel{fig:original-learned-comparison-all}
  \vspace{-4mm}
\end{figure}

\textbf{Embedding distance calibration}~~As described earlier, our
approach to learning an appropriate similarity metric for visual
loop-closure recognition affords a probabilistic interpretation of the
matches proposed. These accurate measures of confidence can be later used to
incorporate these measurements into the back-end pose graph
optimization. Figure~\secref{fig:original-learned-comparison-hists}
illustrates the interpretability of the proposed learned embedding
distance compared to the original feature embedding distance. The
histograms for the $L_2$ embedding distance separation is illustrated for both
positive (in green) and negative (in blue) pairs of features. Here, a
positive pair refers to feature descriptions of images taken at
identical locations, while the negative pairs refer to pairs of
feature descriptions that were taken from at least 50 meters apart
from each other. The figure clearly illustrates how the learned
embedding (\textit{Ours-fc7}) is able to tease apart positive pairs, from those between the negative pairs
of features, enabling an improved classifier (with a more obvious
separator) for place-recognition. Intuitively, the histogram overlap
between the positive and negative probability masses measures the
ambiguity in loop-closure identification, with our learned feature
embedding (\textit{Ours-fc7}) demonstrating the least amount of
overlap.

\textbf{Nearest-Neighbor search in the learned feature embedding
space}~~Once the distances are calibrated in the feature embedding
space, even a na\"ive fixed-radius nearest neighbor strategy, that we
shall refer to as $\varepsilon$-NN, can be surprisingly powerful. In
Figure~\secref{fig:original-learned-comparison-enn}, we show that our
approach is able to achieve high-recall, with considerably strong
precision performance for features that lie within distance $\alpha$
(contrastive loss margin as described in
Section~\secref{subsec:distance-learning}) from each other. 

\begin{figure}[!t]
  \centering 
  {\renewcommand{\arraystretch}{0.6} %
    {\setlength{\tabcolsep}{0.2mm}
      \begin{tabular}{cc}

        \hspace{2mm}\myfiguretitle{Precision-Recall}{7}
        & \hspace{2mm}\myfiguretitle{Recall for increasing $L_2$
          distance}{7} \\
        \includegraphics[width=0.5\columnwidth]{\SECPROJECTDIR/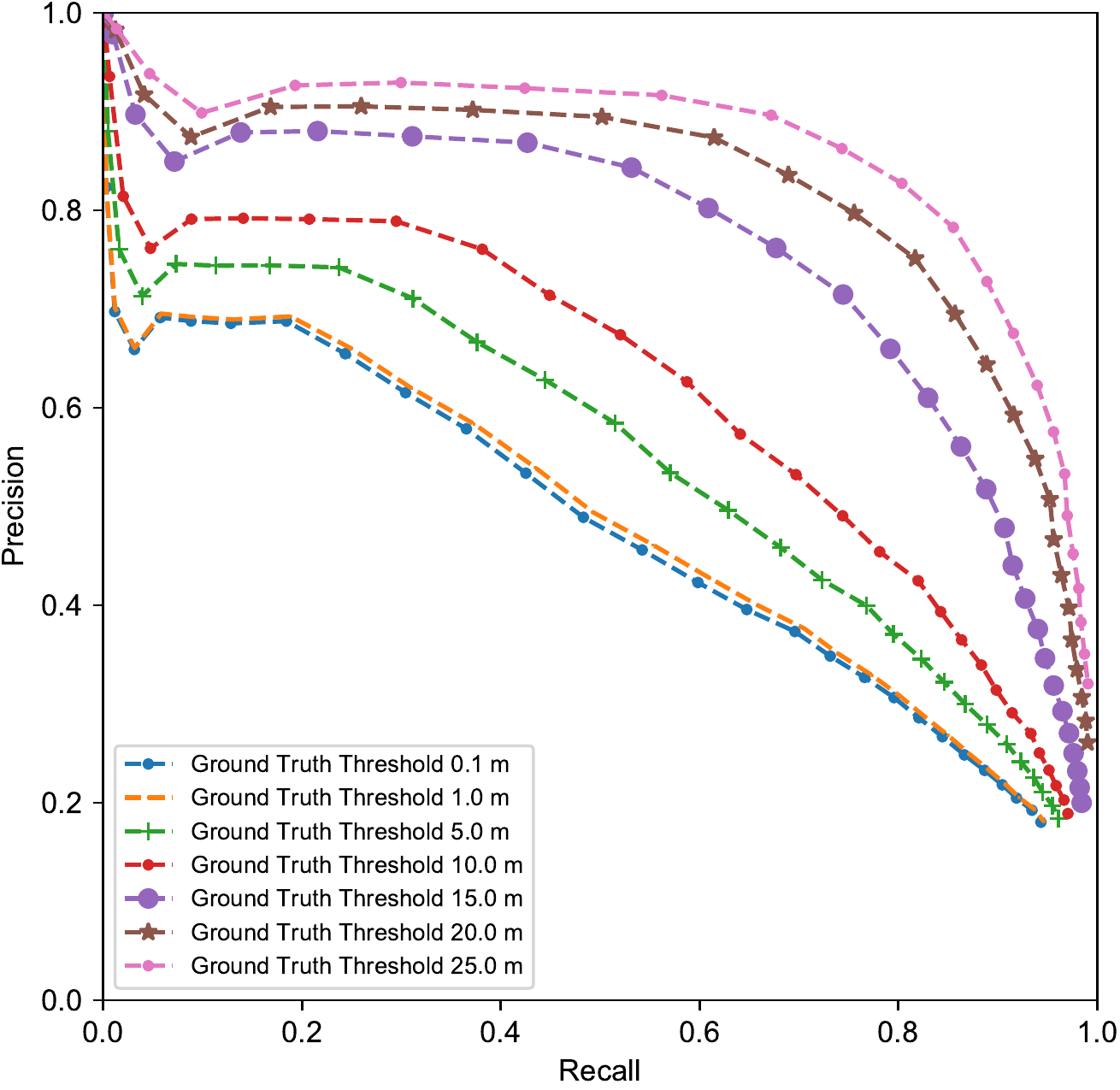}&
        \includegraphics[width=0.5\columnwidth]{\SECPROJECTDIR/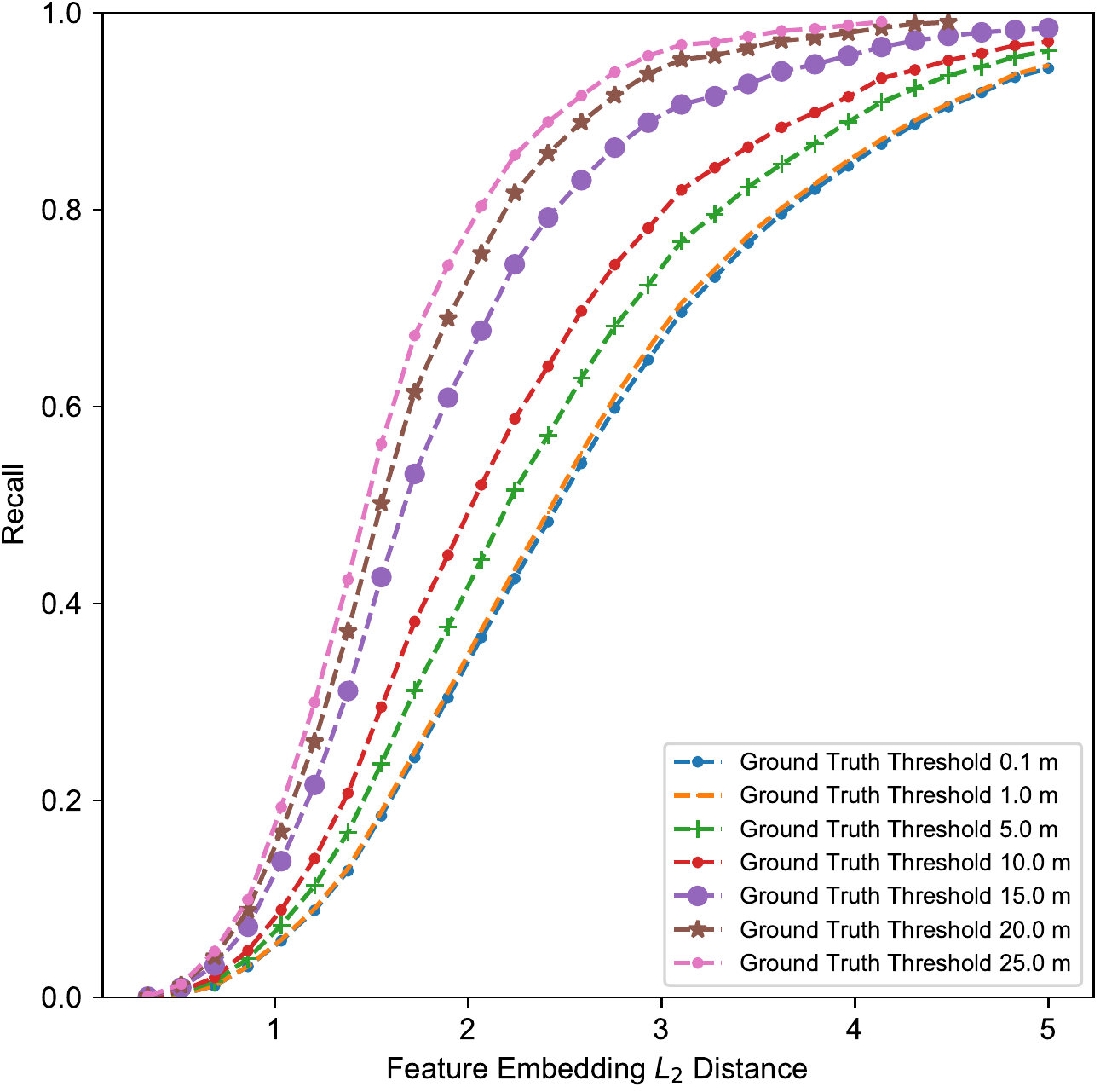}\\
      \end{tabular}}}
  \mycaption{Precision-Recall (\textit{Ours-fc7}) performance for loop-closure
      recognition in the original and learned feature embedding space
      using fixed-radius neighborhood search ($\varepsilon$-nn)}{ The first column
    convincingly shows that our learned feature embedding space is able to
    maintain strong Precision-Recall performance by using $\varepsilon$-nn
    (fixed-radius search). The plot on the {second column} shows the
    recall performance with increasing feature embedding $L_2$ distance
    considered for each query sample. The Siamese network was trained with
    a contrastive loss margin of $\alpha=10$, which distorts the embedding
    space such that positive pairs are encouraged to only be separated by an $L_2$
    distance of 10 or lower. The figure on the \textit{right} shows that in the
    learned feature embedding space (\textit{Ours-fc7}), we are able to
    capture most candidate loop-closures within an $L_2$ distance of 5
    from the query sample, as more matching neighbors are considered.}
  \seclabel{fig:original-learned-comparison-enn}
  \vspace{-4mm}
\end{figure}

Furthermore, the feature embedding can also be used in the context of
image retrieval with strong recall performance via na\"ive $k$-Nearest
Neighbor ($k$-NN)
search. Figure~\secref{fig:original-learned-comparison-knn} compares
the precision-recall performance of the $k$-NN strategy on the
original and learned embedding space, and shows a considerable
performance gain in the learned embedding space. Furthermore, the
\textit{recall} performance also tends to be higher for the learned embedding
space as compared to the original descriptor space.

\begin{figure}[!t]
  \centering 
  {\renewcommand{\arraystretch}{0.2} %
    {\setlength{\tabcolsep}{0.2mm}
      \begin{tabular}{cc}
        \hspace{2mm}\myfiguretitle{Precision-Recall (\textit{Ours-fc7})}{6}
        & \hspace{2mm}\myfiguretitle{Recall with increasing k-NN (\textit{Ours-fc7})}{6} \\
        \includegraphics[width=0.48\columnwidth]{\SECPROJECTDIR/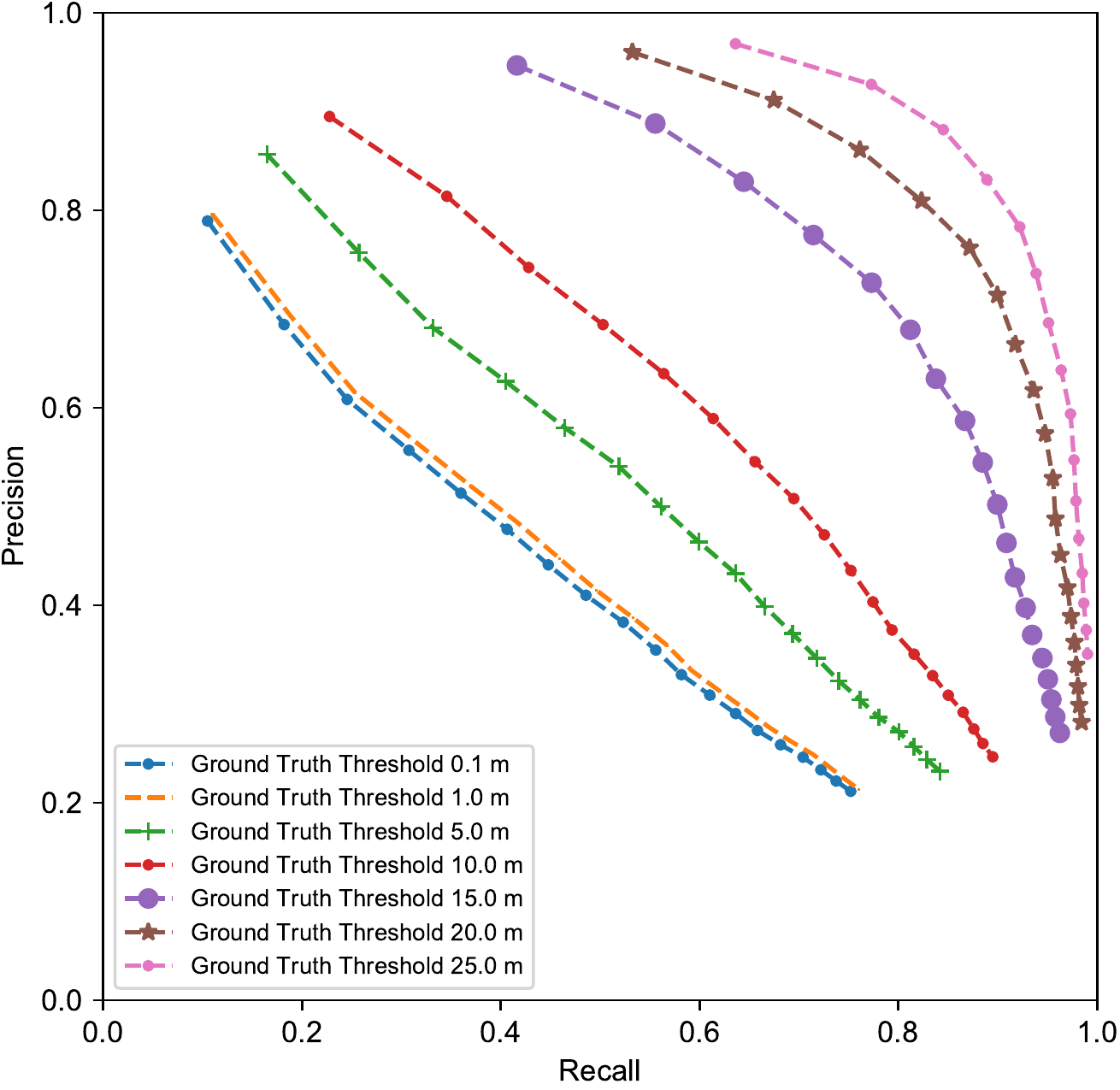}&
        \includegraphics[width=0.48\columnwidth]{\SECPROJECTDIR/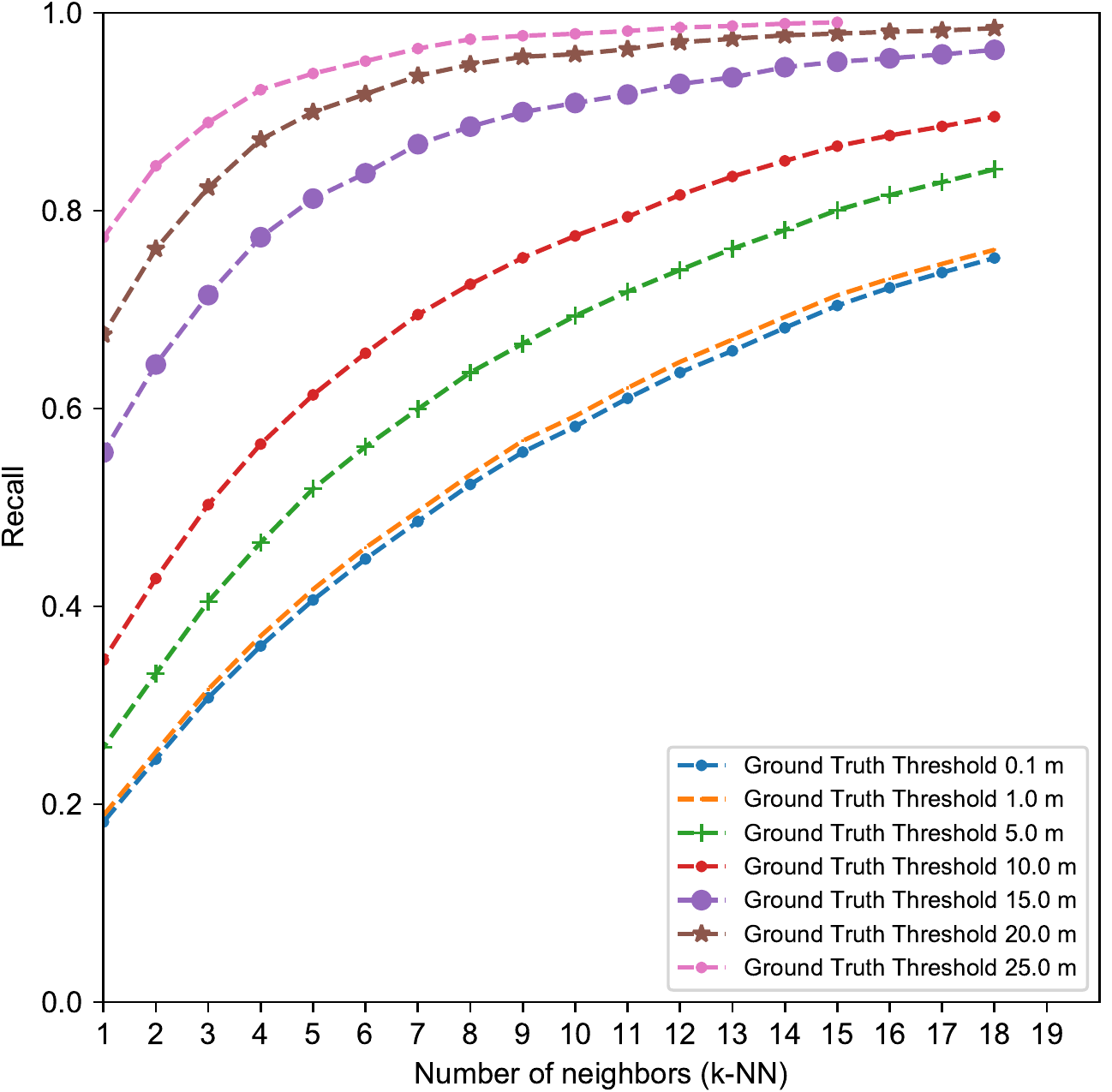}\vspace{2mm}\\

        \hspace{2mm}\myfiguretitle{Precision-Recall (\textit{Places365-fc7})}{6}
        & \hspace{2mm}\myfiguretitle{Recall with increasing k-NN (\textit{Places365-fc7})}{6} \\
        \includegraphics[width=0.48\columnwidth]{\SECPROJECTDIR/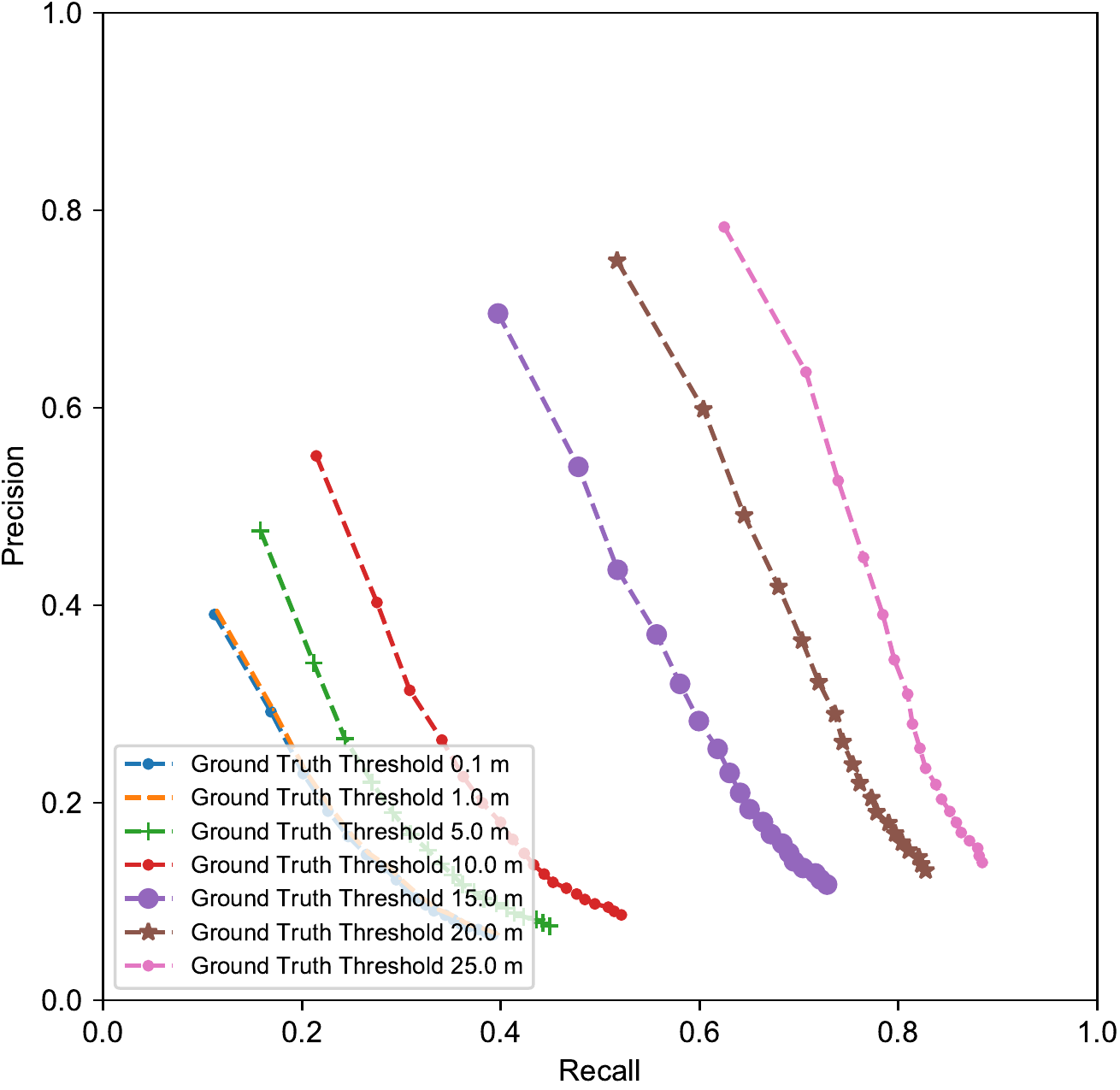}&
        \includegraphics[width=0.48\columnwidth]{\SECPROJECTDIR/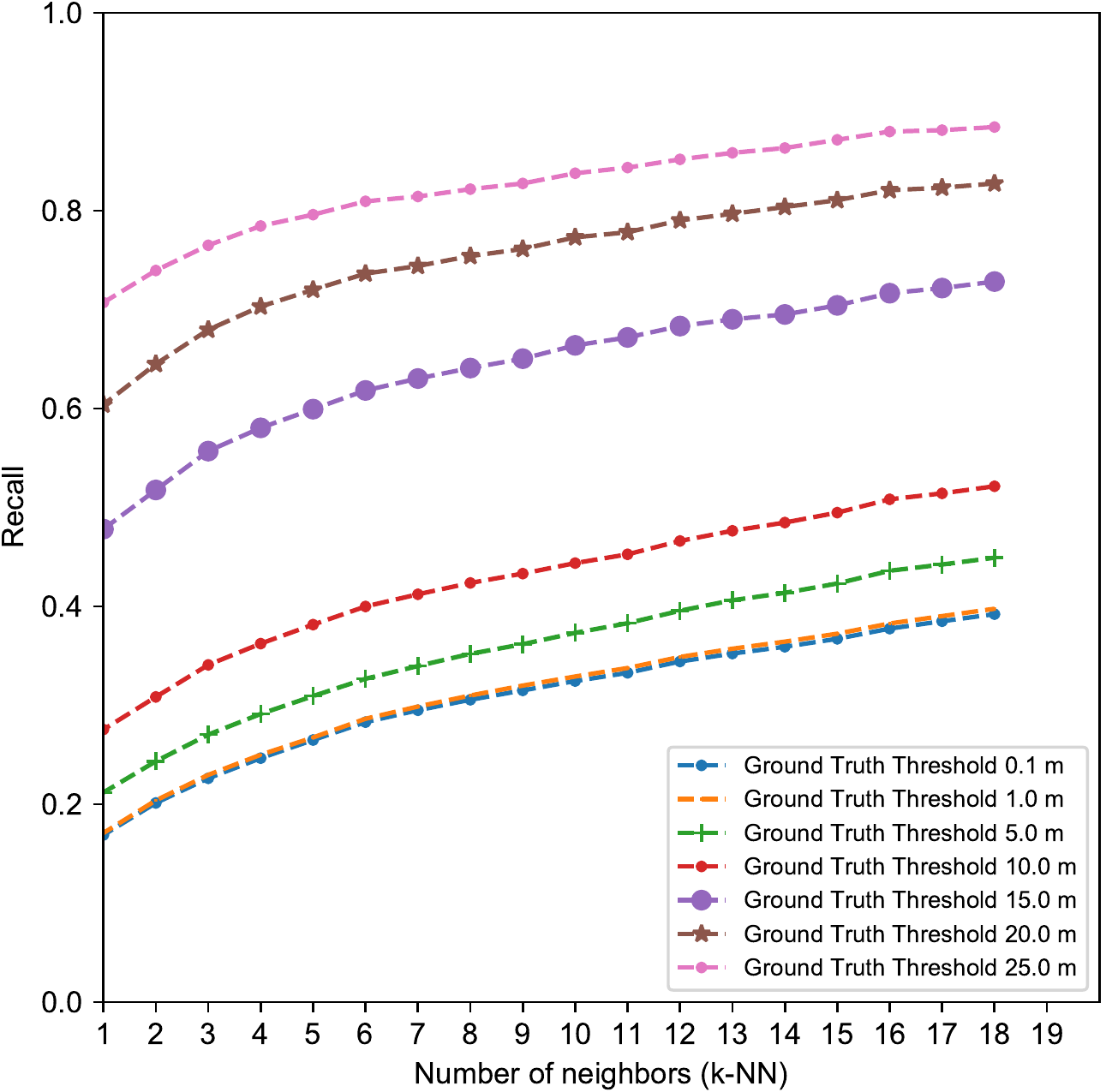}\vspace{2mm}\\
      \end{tabular}}}
  \mycaption{Precision-Recall performance for loop-closure
      recognition in the original and learned feature embedding space using
      k-Nearest Neighbors}{ The first column shows that our
    learned feature embedding space is able to perform considerably better than the pre-trained layers
    (Places365-AlexNet \textit{fc7}). The plot on the
    {second column} shows the recall performance with increasing set of
    neighbors considered for each query sample. Using the learned feature
    embedding space (\text{Ours-fc7}), we are able to capture more
    candidate loop-closures within the closest 20 neighbors of the query
    sample.}
  \seclabel{fig:original-learned-comparison-knn}
  \vspace{-4mm}
\end{figure}

\begin{figure*}[!t]
  \centering 
  {\renewcommand{\arraystretch}{0.6} %
    {\setlength{\tabcolsep}{0.2mm}
      \begin{tabular}{cccc}

        \hspace{4mm}\myfiguretitle{Separation distance histogram (\textit{Ours-fc7})}{5}
        &\hspace{4mm}\myfiguretitle{Separation distance histogram (\textit{conv3})}{5}
        &\hspace{4mm}\myfiguretitle{Separation distance histogram (\textit{conv4})}{5}
        &\hspace{4mm}\myfiguretitle{Separation distance histogram (\textit{conv5})}{5}\\
        
        \includegraphics[scale=0.32]{\SECPROJECTDIR/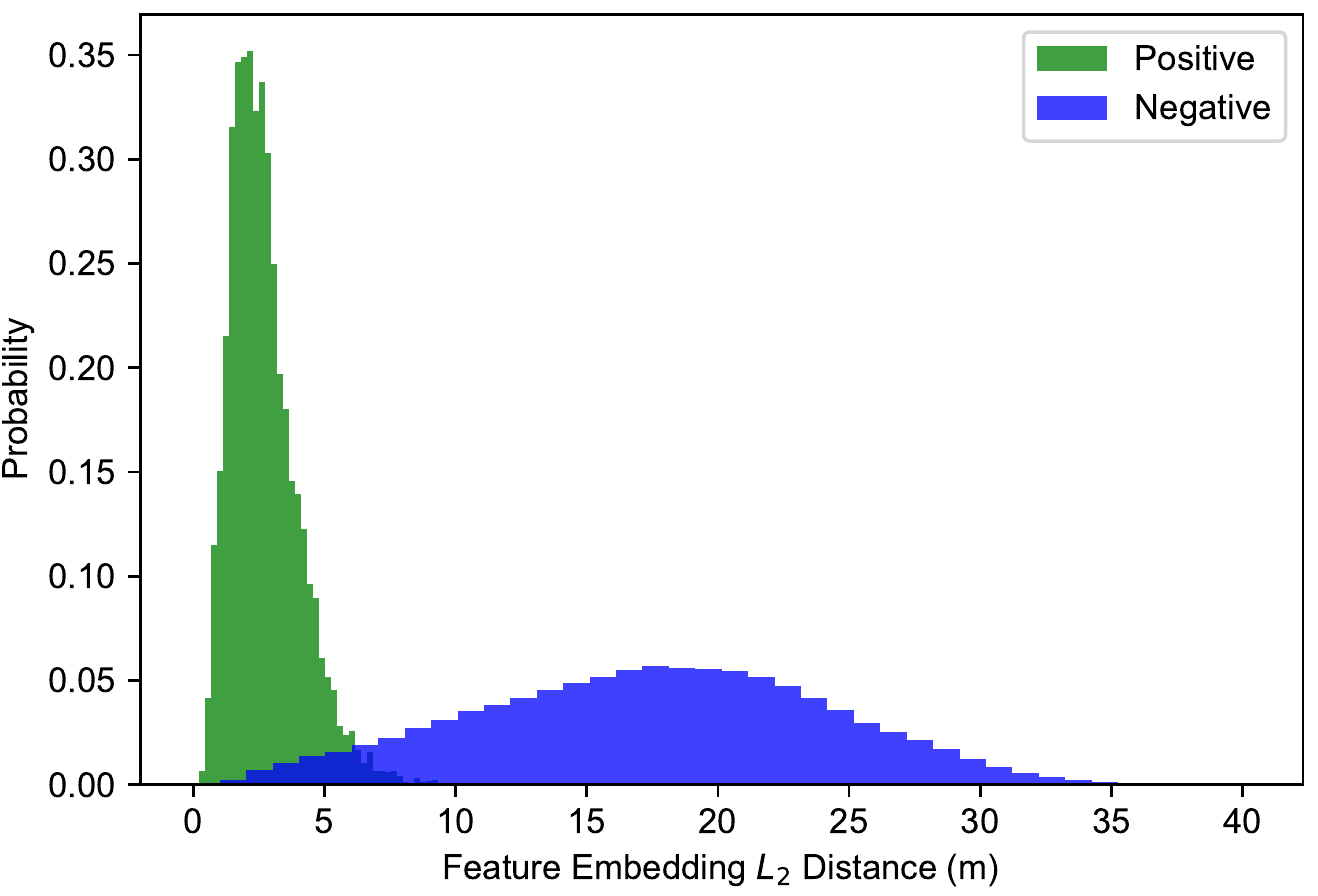}
        &\includegraphics[scale=0.32]{\SECPROJECTDIR/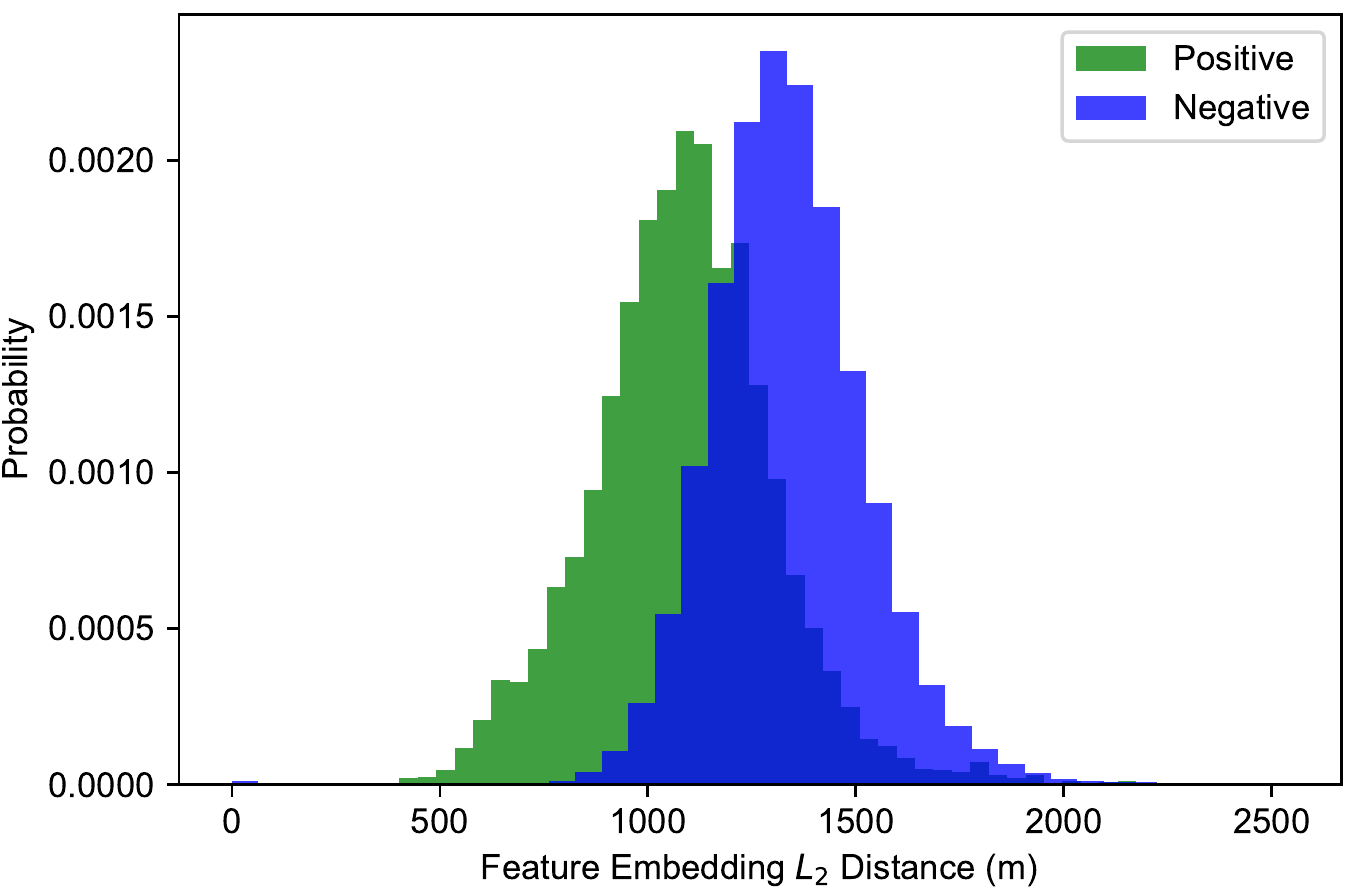}
        &\includegraphics[scale=0.32]{\SECPROJECTDIR/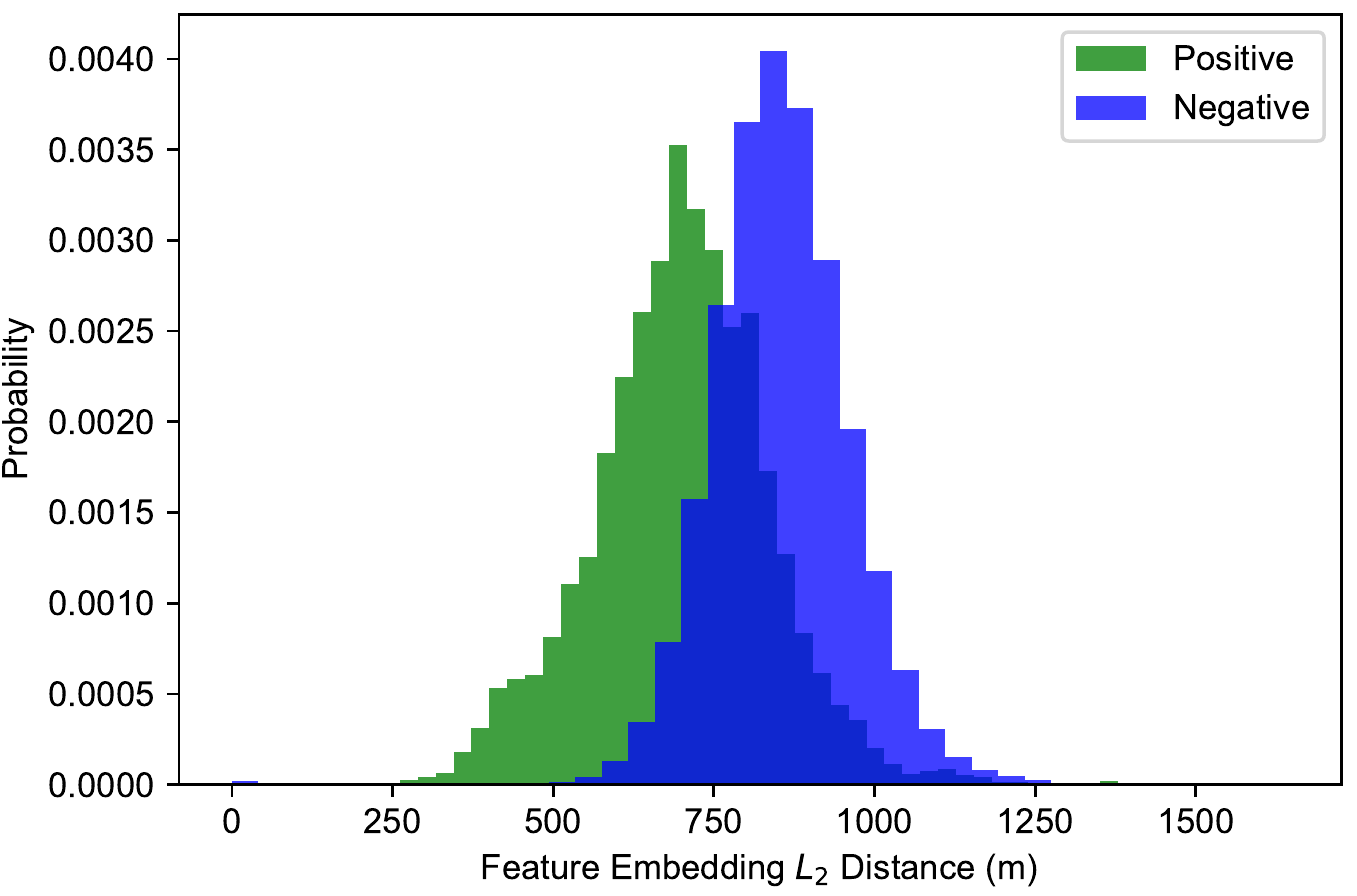}
        &\includegraphics[scale=0.32]{\SECPROJECTDIR/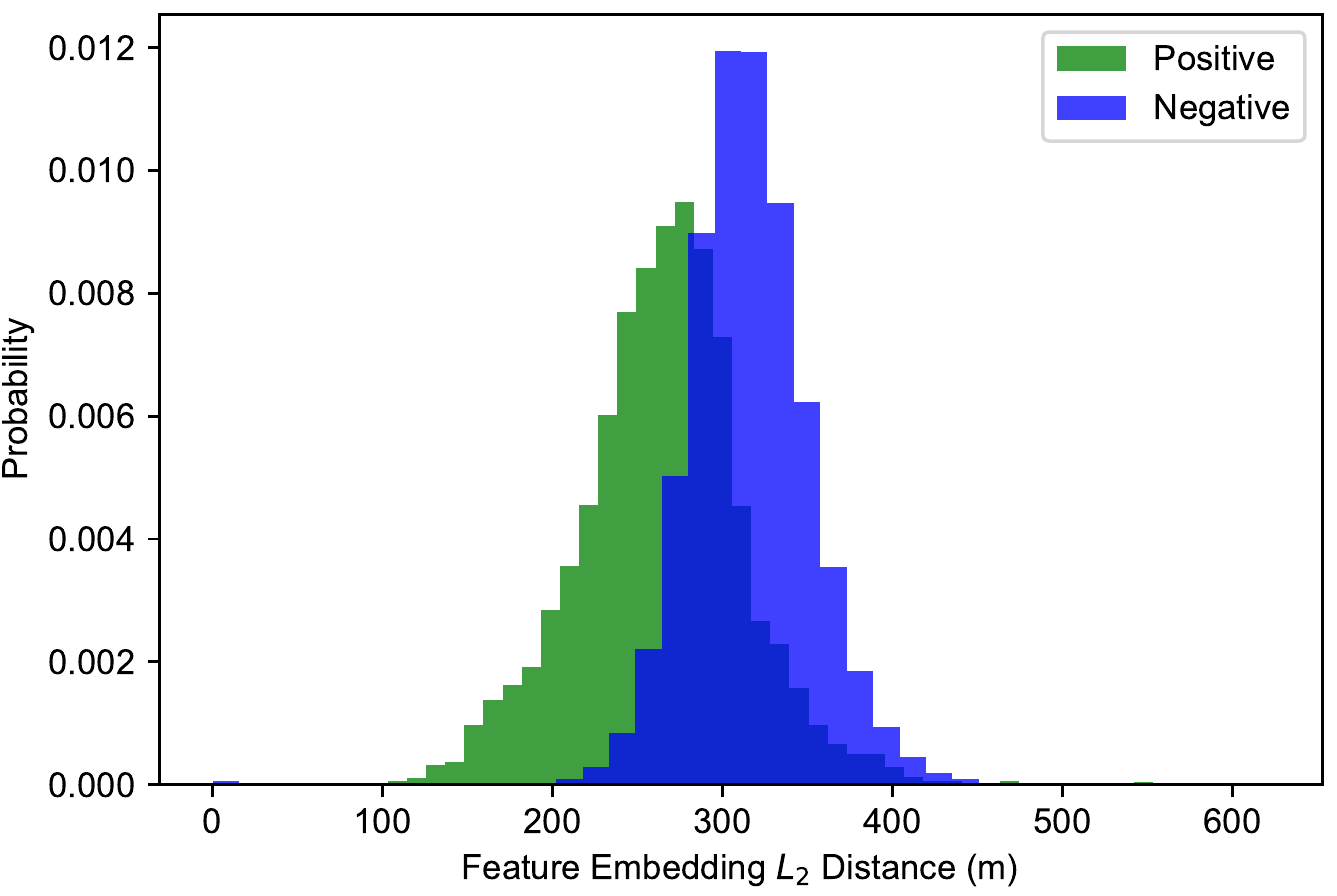}\\

        \hspace{4mm}\myfiguretitle{Separation distance histogram (\textit{pool5})}{5}
        &\hspace{4mm}\myfiguretitle{Separation distance histogram (\textit{fc6})}{5}
        &\hspace{4mm}\myfiguretitle{Separation distance histogram (\textit{fc7)}}{5}
        &\hspace{4mm}\myfiguretitle{Separation distance histogram (\textit{fc8})}{5}\\

        \includegraphics[scale=0.32]{\SECPROJECTDIR/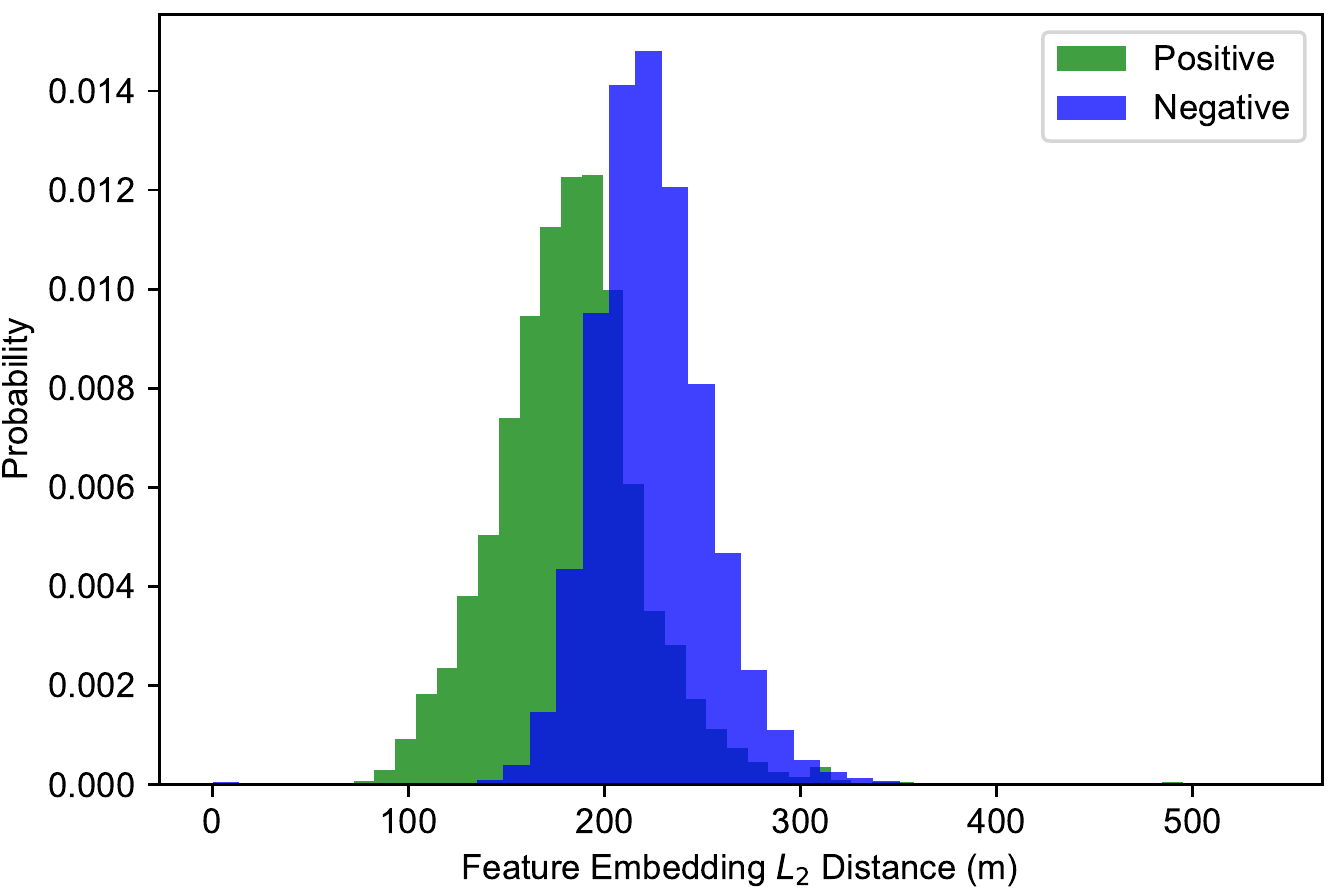}
        &\includegraphics[scale=0.32]{\SECPROJECTDIR/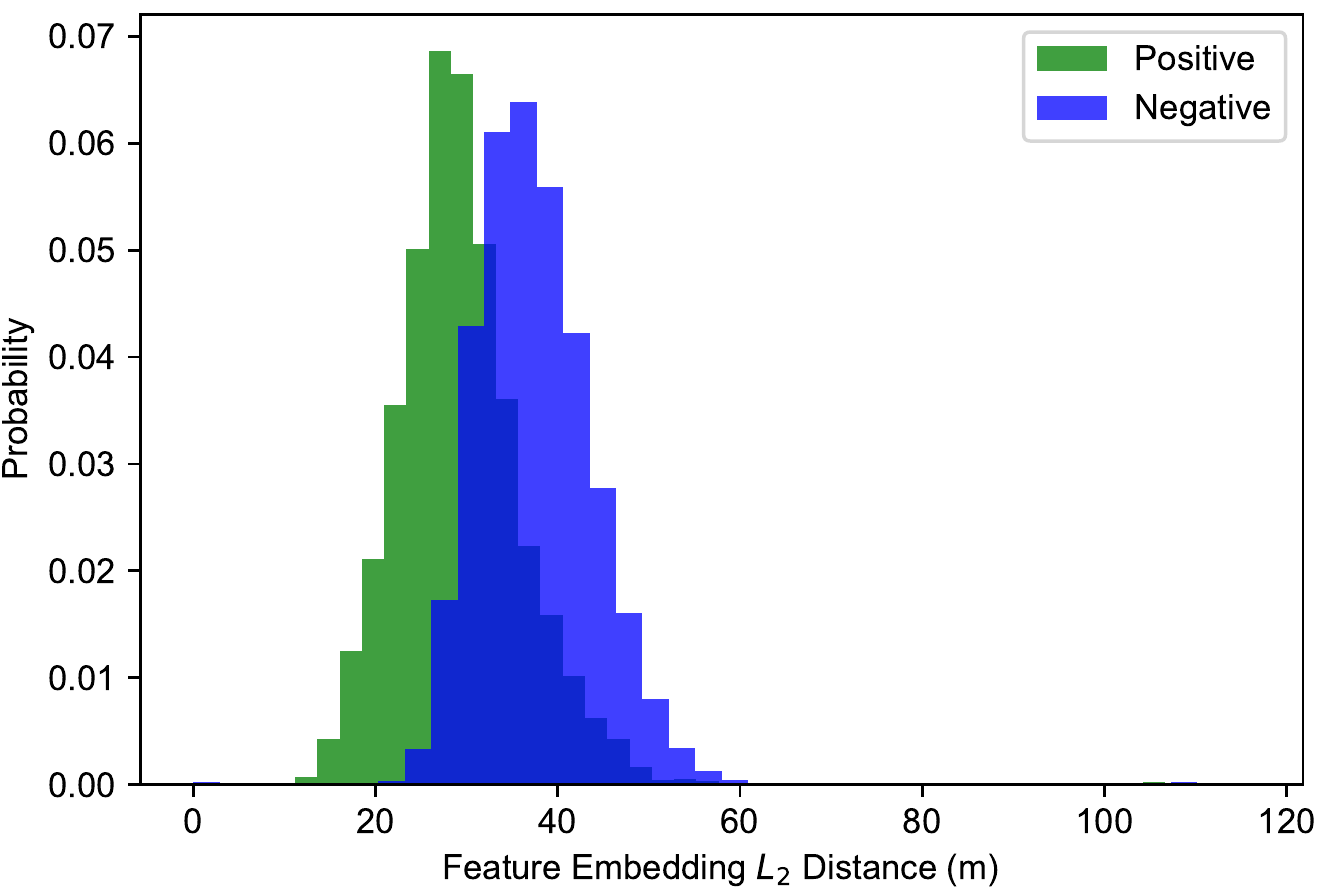}
        &\includegraphics[scale=0.32]{\SECPROJECTDIR/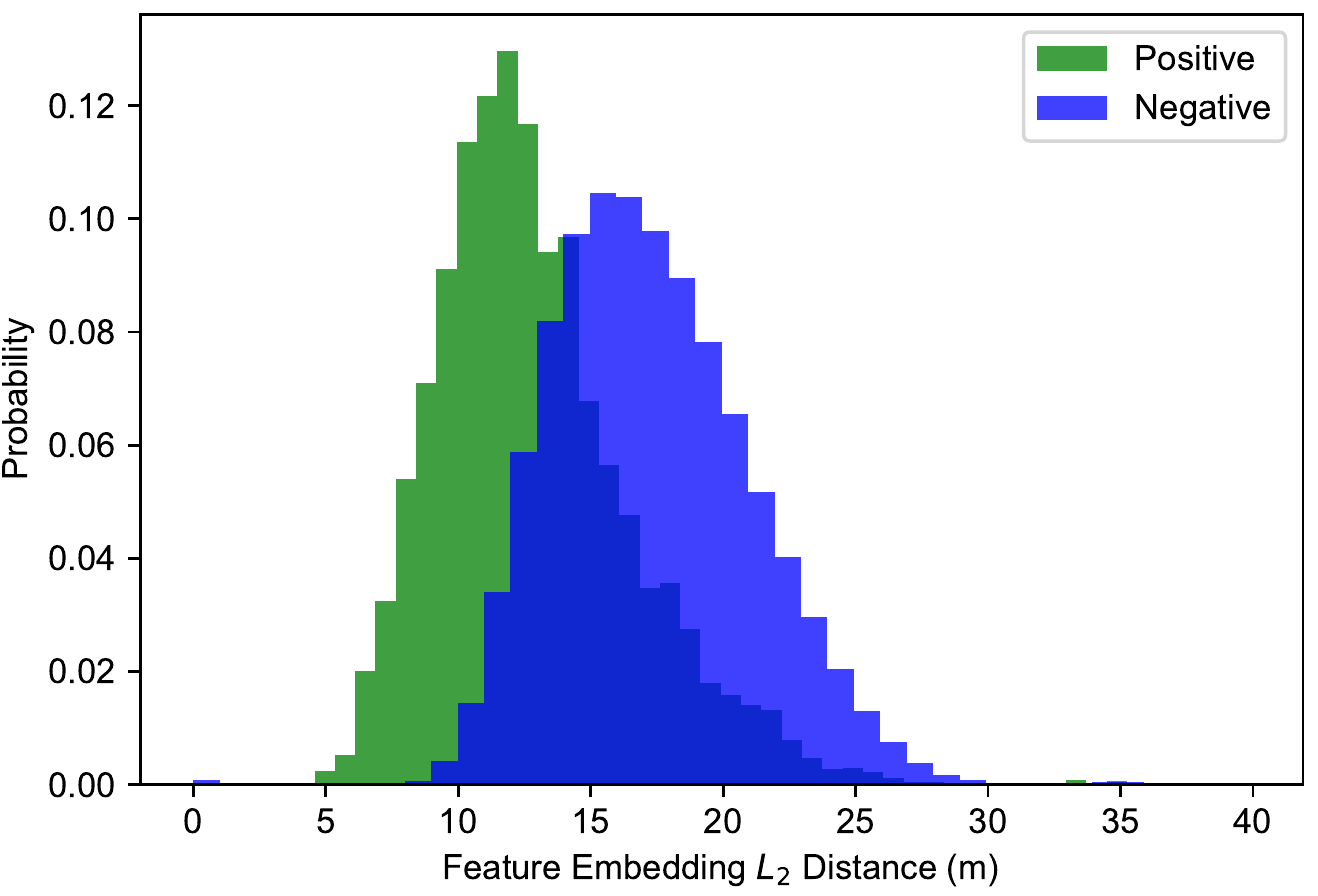}
        &\includegraphics[scale=0.32]{\SECPROJECTDIR/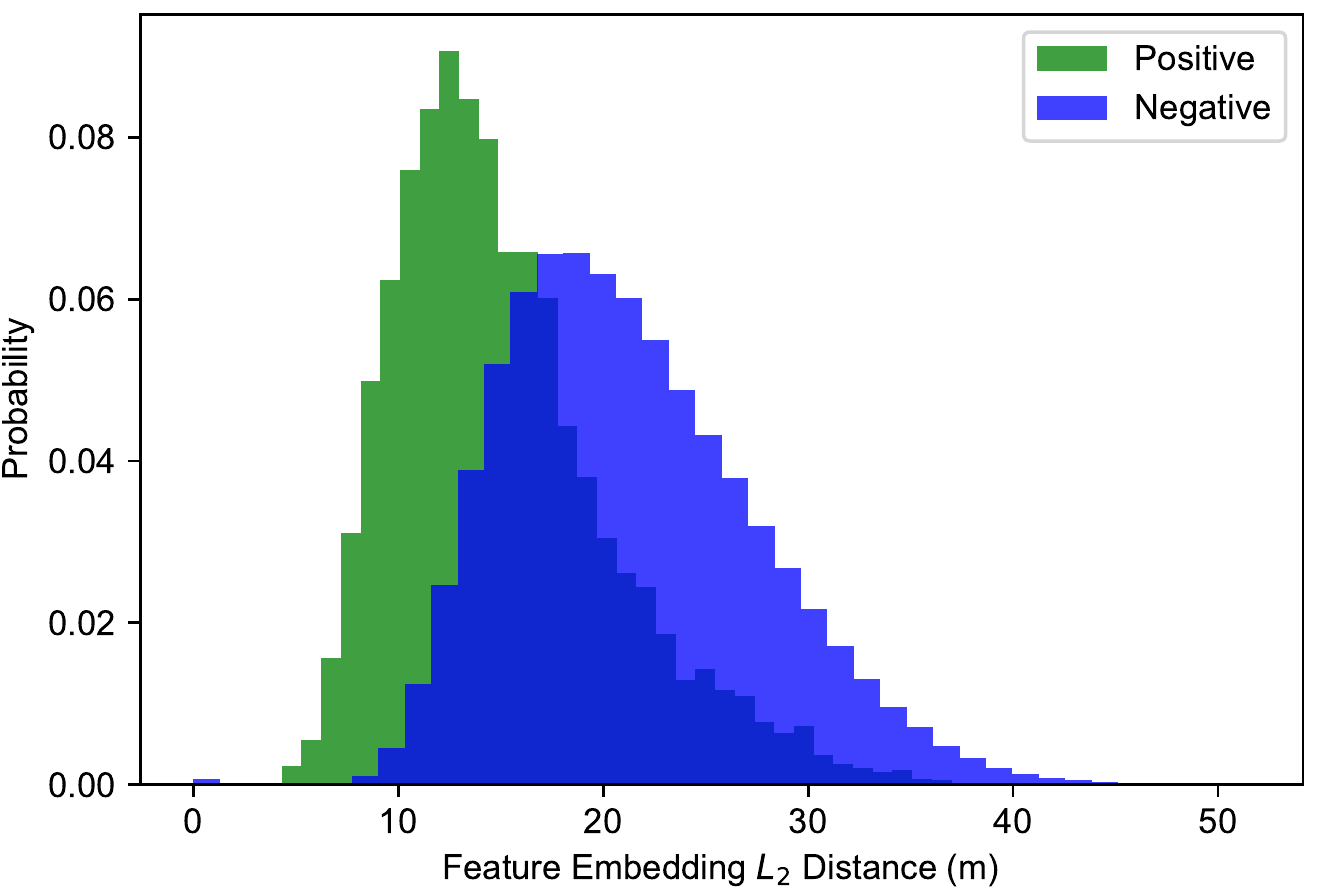}\\
        
      \end{tabular}}}
  \mycaption{Separation distance calibration}{ The histograms of
    $L_2$ distances between positive and negative examples are shown
    for the various feature descriptions with the pre-trained
    Places365-AlexNet model. Our learned model is able to fine-tune
    intermediate layers and distort the feature embedding such that
    the distances between positive and negative examples (similar and
    dissimilar places) are well-calibrated. This is seen especially in
    the first plot (top row, far left~\textit{Ours-fc7}), where the probability mass for
    positive and negative examples are better separated with reduced
    overlap, while the other histograms are not well-separated in the
    feature embedding space.}
  \seclabel{fig:original-learned-comparison-hists}
  \vspace{-2mm}
\end{figure*}

\subsection{Localization performance within visual-SLAM front-ends}

Figure~\secref{fig:slam-with-learned-localization} shows
the trajectory of the optimized pose-graph leveraging the constraints
proposed by our learned loop-closure proposal method. The visual
place-recognition module determines constraints between temporally
distant nodes in the pose-graph that are likely to be associated with
the same physical location. To evaluate the localization module
independently, we simulate drift in the odometry chains by injecting
noise in the individual ground truth odometry measurements.

\begin{figure}[h]
  \centering 
  {\renewcommand{\arraystretch}{0.6} %
    {\setlength{\tabcolsep}{0.1mm}
      \begin{tabular}{M{1.5mm}
        M{0.24\columnwidth}
        M{0.24\columnwidth}
        M{0.24\columnwidth}
        M{0.24\columnwidth}}

        \multicolumn{5}{c}{\footnotesize \textbf{KITTI Sequence 00}} \\\\
        \rotatebox[x=0mm]{90}{\scriptsize Measured}& 
        \includegraphics[width=0.24\columnwidth,clip,trim=35mm 5mm 40mm 155mm]{\SECPROJECTDIR/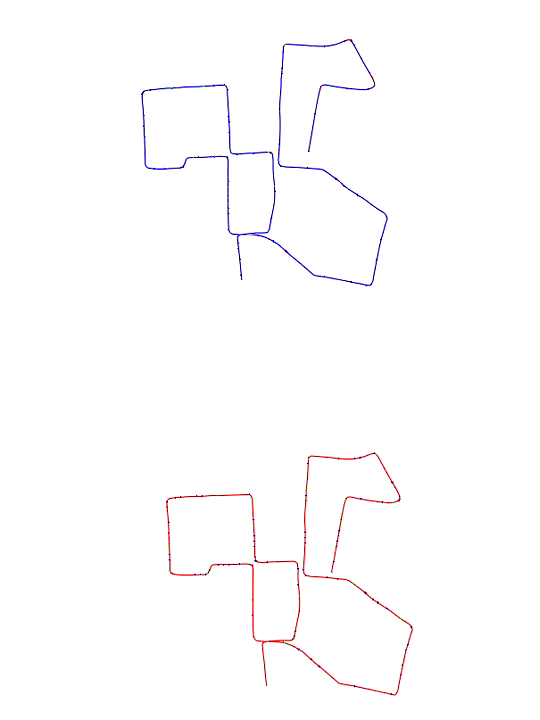}
        &\includegraphics[width=0.24\columnwidth,clip,trim=35mm 5mm 40mm 155mm]{\SECPROJECTDIR/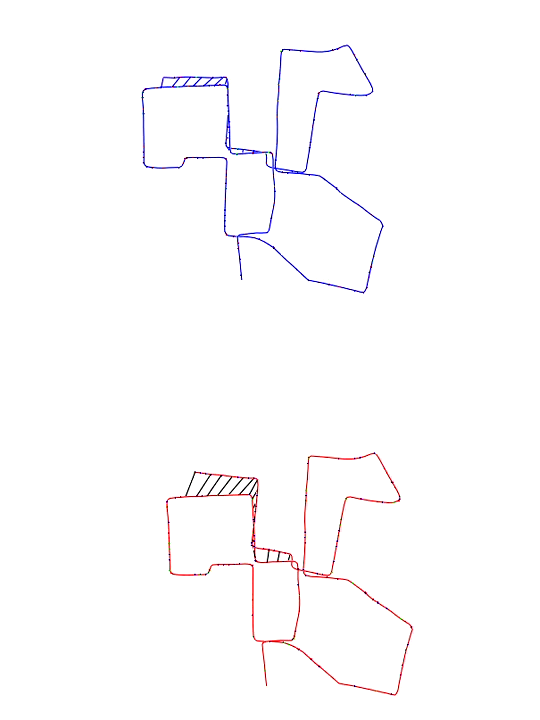}
        &\includegraphics[width=0.24\columnwidth,clip,trim=35mm 5mm 40mm 155mm]{\SECPROJECTDIR/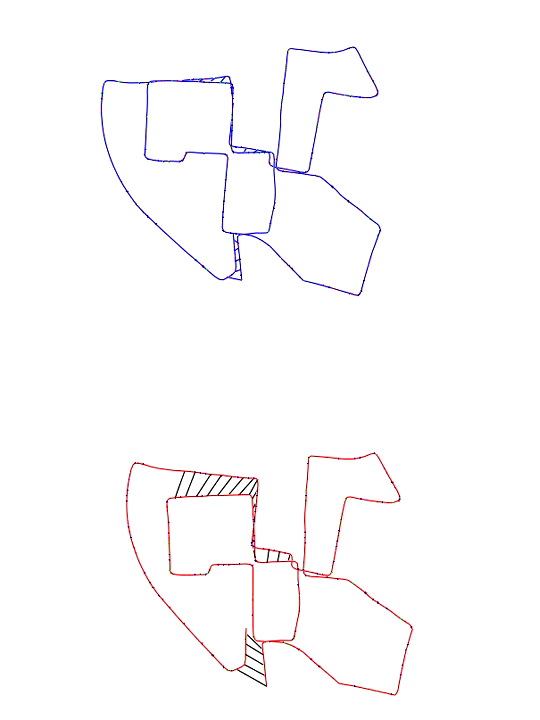}
        &\includegraphics[width=0.24\columnwidth,clip,trim=35mm 5mm 40mm 155mm]{\SECPROJECTDIR/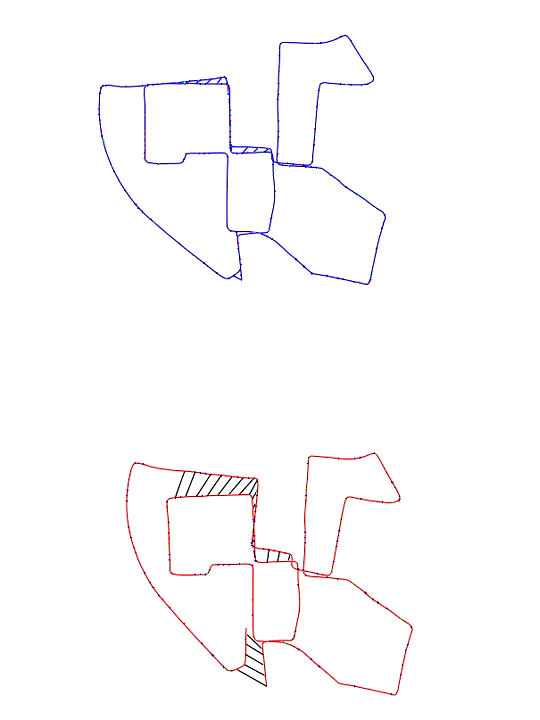}\\

        \rotatebox[x=0mm]{90}{\scriptsize Optimized}& 
        \includegraphics[width=0.24\columnwidth,clip,trim=25mm 148mm 40mm 5mm]{\SECPROJECTDIR/results/slam_with_localization/kitti-1.png}
        &\includegraphics[width=0.24\columnwidth,clip,trim=25mm 148mm 40mm 5mm]{\SECPROJECTDIR/results/slam_with_localization/kitti-2.png}
        &\includegraphics[width=0.24\columnwidth,clip,trim=25mm 148mm 40mm 5mm]{\SECPROJECTDIR/results/slam_with_localization/kitti-3.png}
        &\includegraphics[width=0.24\columnwidth,clip,trim=25mm 148mm 40mm 5mm]{\SECPROJECTDIR/results/slam_with_localization/kitti-4.png}\\
        & $t_1$ & $t_2$ & $t_3$ & $T$\\    

        \\\\
        \multicolumn{5}{c}{\footnotesize \textbf{St. Lucia Dataset}} \\\\
        \rotatebox[x=0mm]{90}{\scriptsize Measured}& 
        \includegraphics
        [width=0.24\columnwidth,clip,trim=0 20mm 20mm 120mm]{\SECPROJECTDIR/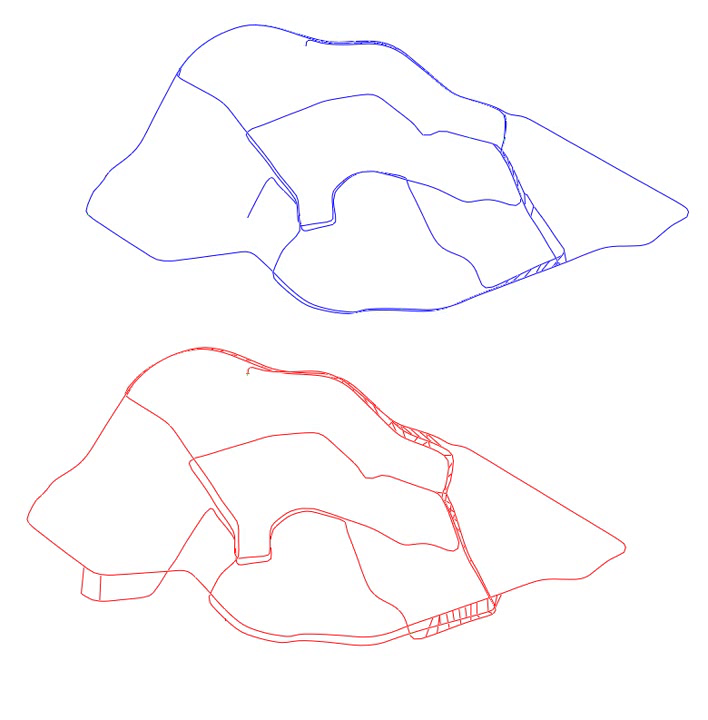}
          &\includegraphics
            [width=0.24\columnwidth,clip,trim=0 20mm 20mm 120mm]{\SECPROJECTDIR/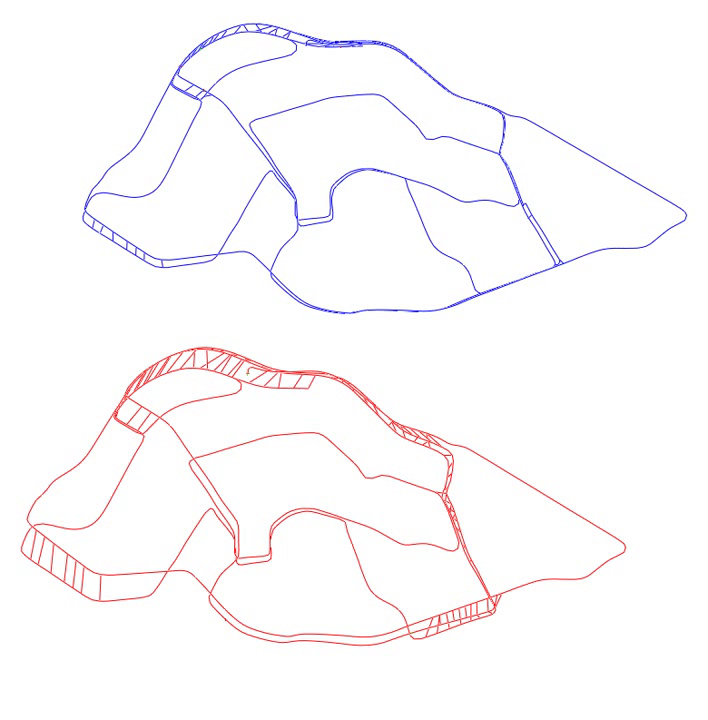}
          &\includegraphics
            [width=0.24\columnwidth,clip,trim=0 20mm 20mm 120mm]{\SECPROJECTDIR/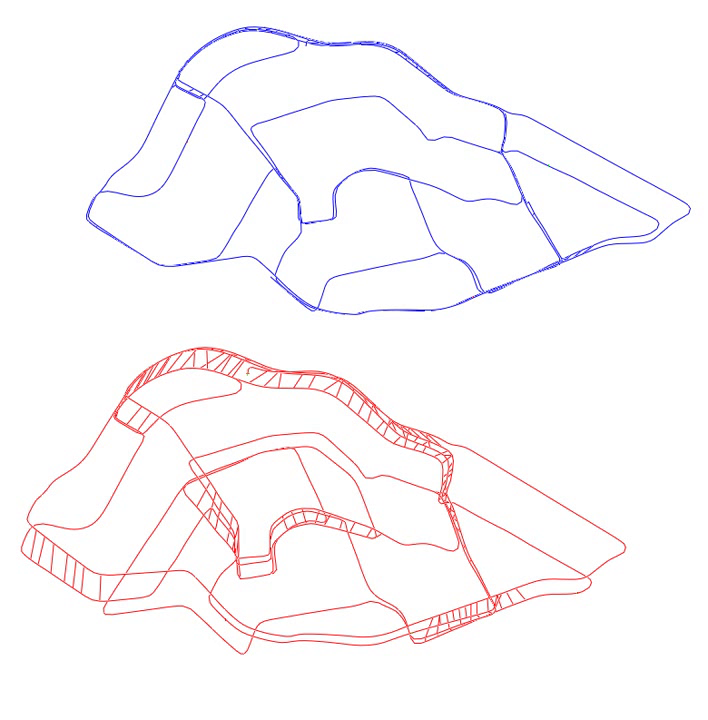}
          &\includegraphics
            [width=0.24\columnwidth,clip,trim=0 20mm 20mm 120mm]{\SECPROJECTDIR/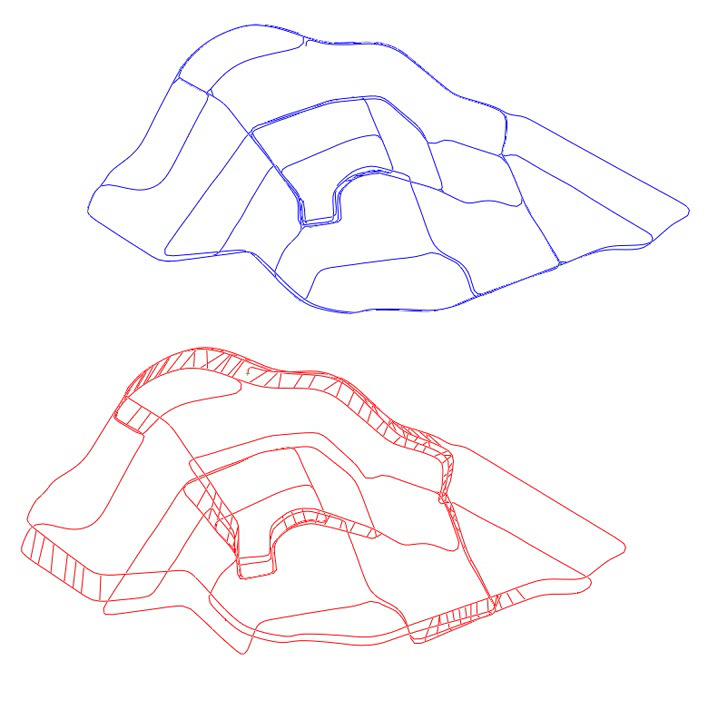}\\

        \rotatebox[x=0mm]{90}{\scriptsize Optimized}& 
        \includegraphics
        [width=0.24\columnwidth,clip,trim=20mm 135mm 0 0]{\SECPROJECTDIR/results/slam_with_localization/st_lucia_switch-1.png}
          &\includegraphics
            [width=0.24\columnwidth,clip,trim=20mm 135mm 0 0]{\SECPROJECTDIR/results/slam_with_localization/st_lucia_switch-2.png}
          &\includegraphics
            [width=0.24\columnwidth,clip,trim=20mm 135mm 0 0]{\SECPROJECTDIR/results/slam_with_localization/st_lucia_switch-3.png}
          &\includegraphics
            [width=0.24\columnwidth,clip,trim=20mm 135mm 0 0]{\SECPROJECTDIR/results/slam_with_localization/st_lucia_switch-4.png}\\
        & $t_1$ & $t_2$ & $t_3$ & $T$\\            
      \end{tabular}}}\vspace{2mm}
  \mycaption{Vision-based Pose-Graph SLAM with our learned
      place-recognition module}{ The two sets of plots show the
    measured ({\color{fullred} in
    red}) and optimized ({\color{navyblue}in blue}) pose-graph for a
    particular KITTI and St Lucia session. The crossed edges in the
    measured pose-graph corresponds to loop-closure candidates proposed by
    our learned place-recognition module. As more measurements are added
    and loop-closures are proposed ($t_1 < t_2 < t_3 < T$), the pose-graph
    optimization accurately recovers the true
    trajectory of the vehicle across the entire session. For both
    sessions, we inject odometry noise to simulate drift in typical
    odometry estimates.}
  \vspace{-8mm}
  \seclabel{fig:slam-with-learned-localization}
\end{figure}

The trajectory recovered from sequential noisy odometry measurements
are shown in red, as more measurements are added ($t_1 < t_2 <
t_3 < T$). With every new image, the image is mapped
into the appropriate embedding space and subsequently used to query the
database for a potential loop-closure. The loop-closures are realized as weak zero
rotation and translation relative pose-constraints connecting the
query node and the matched node. The recovered trajectories after the
pose-graph optimization (in blue) shows consistent
long-range, and drift-free trajectories that the vehicle traversed.

\subsection{Implementation details}
\seclabel{subsec:implementation-details}

\textbf{Network and Training} We take the pre-trained Places205
AlexNet~\cite{zhou2014learning,zhou2016places}, and set all the layers
before and including \textit{pool5} layer to be fixed, while the rest
of the fully-connected layers (\textit{fc6,~fc7}) are allowed to be fine-tuned. The
resulting network is used as a base network to construct the
Siamese Network with shared weights (See
Section~\secref{subsec:distance-learning}).  We follow the
distance-weighted sampling scheme as proposed by~\citet{1706.07567},
and sample 10 times more negative examples as positive examples. The
class weights are scaled appropriately to avoid any class imbalance
during training. In all our experiments, we set the sampling threshold
$\tau^{\BR\Bt}$ to 0.9, that ensures that identical places have
considerable overlap in their viewing frustums. We train the model for
3000 epochs, with each epoch roughly taking 10s on an NVIDIA Titan X
GPU. For most datasets including KITTI and St. Lucia Dataset, we train on 3-5 data
sessions collected from the vehicle, and test on a completely new
session. 

\textbf{Pose-Graph Construction and Optimization} We use
GTSAM\footnote{\scriptsize\url{http://collab.cc.gatech.edu/borg/gtsam}}
to construct the pose-graph and establish loop-closure constraints for
pose-graph optimization. For validating the loop-closure recognition
module, the odometry constraints are recovered from the ground truth,
with noise injected to simulate dead-reckoned drift in the odometry
estimate. They are incorporated as a relative-pose constraint
parametrized in $SE(2)$ with $\expnumber{1}{-3}$~rad rotational noise
and $\expnumber{5}{-2}$~m translation noise. We incorporate the
loop-closure constraints as zero translation and rotation
relative-pose constraint with a weak translation and
rotation covariance of $3$~m and $0.3$~rad respectively. The
constraints are incrementally added and solved using
iSAM2~\cite{kaess2012isam2} as the measurements are recovered.

\section{Conclusion} In this work, we develop a
\textit{self-supervised} approach to place recognition in robots. By
leveraging the synchronization between sensors, we propose a method to
transfer and learn a metric for image-image similarity in an embedded
space by sampling corresponding information from a GPS-aided
navigation solution. Through experiments, we show that the newly
learned embedding can be particularly powerful for the task of visual
place-recognition as the embedded distances are well-calibrated for
efficient indexing and accurate retrieval. We believe that such
techniques can be especially powerful as the robot can quickly
fine-tune their pre-trained models to their operating environments,
by simply collecting more relevant experiences.

{
  \footnotesize
  \bibliographystyle{\BIBSTYLE}
  \bibliography{tex/references}
}

\end{document}